\documentclass[10pt,twocolumn,letterpaper]{article}

\usepackage{cvpr}
\usepackage{times}
\usepackage{epsfig}
\usepackage{graphicx}
\usepackage{amsmath}
\usepackage{amssymb}

\usepackage{siunitx}
\usepackage{booktabs}
\usepackage{subcaption}
\usepackage{paralist}
\usepackage{pifont}
\usepackage[sort,nocompress]{cite}

\usepackage[pagebackref=true,breaklinks=true,colorlinks,bookmarks=false]{hyperref}

\DeclareGraphicsExtensions{.pdf,.jpg,.png}
\graphicspath{{figs/}}

\newcommand{\secref}[1]{Section~\ref{sec:#1}}
\newcommand{\figref}[1]{Figure~\ref{fig:#1}}
\newcommand{\tblref}[1]{Table~\ref{tbl:#1}}

\cvprfinalcopy % *** Uncomment this line for the final submission

\def\cvprPaperID{2487} % *** Enter the CVPR Paper ID here

\ifcvprfinal\fi
\begin{document}

\title{Dynamic Traffic Modeling from Overhead Imagery}

\author{
  \centering
  \begin{minipage}{.9\linewidth}
    \centering
    \begin{minipage}{2.0in}
      \centering
      Scott Workman
    \end{minipage}
    \begin{minipage}{2.0in}
      \centering
      Nathan Jacobs
    \end{minipage}
    \\[.2cm]
    \begin{minipage}{2.0in}
      \centering
      DZYNE Technologies
    \end{minipage}
    \begin{minipage}{2.0in}
      \centering
      University of Kentucky
    \end{minipage}
  \end{minipage}
}

\maketitle

\begin{abstract}
    Our goal is to use overhead imagery to understand patterns in
    traffic flow, for instance answering questions such as how fast
    could you traverse Times Square at 3am on a Sunday. A traditional
    approach for solving this problem would be to model the speed of
    each road segment as a function of time. However, this strategy is
    limited in that a significant amount of data must first be
    collected before a model can be used and it fails to generalize to
    new areas. Instead, we propose an automatic approach for
    generating dynamic maps of traffic speeds using convolutional
    neural networks. Our method operates on overhead imagery, is
    conditioned on location and time, and outputs a local motion model
    that captures likely directions of travel and corresponding travel
    speeds. To train our model, we take advantage of historical
    traffic data collected from New York City. Experimental results
    demonstrate that our method can be applied to generate accurate
    city-scale traffic models.
\end{abstract}

\section{Introduction}

Road transportation networks have become extremely large and complex.
According to the Bureau of Transportation
Statistics~\cite{sprung2018transportation}, there are approximately
6.6 million kilometers of roads in the United States alone. For most
individuals, navigating these complex road networks is a daily
challenge. A recent study found that the average driver in the U.S.
travels approximately \num{17500} kilometers per year in their
vehicle, which equates to more than 290 hours behind the
wheel~\cite{triplett2016american}.

As such, traffic modeling and analysis has become an increasingly
important topic for urban development and planning. The Texas A\&M
Transportation Institute~\cite{schrank2019urban} estimated that in
2017, considering 494 U.S. urban areas, there were 8.8 billion
vehicle-hours of delay and 12.5 billion liters of wasted fuel,
resulting in a congestion cost of 179 billion dollars. Given these
far-reaching implications, there is significant interest in
understanding traffic flow and developing new methods to counteract
congestion.

\begin{figure}
    \centering
    \frame{\includegraphics[trim={2in, 2in, 2in, 2in},clip,width=.49\linewidth]{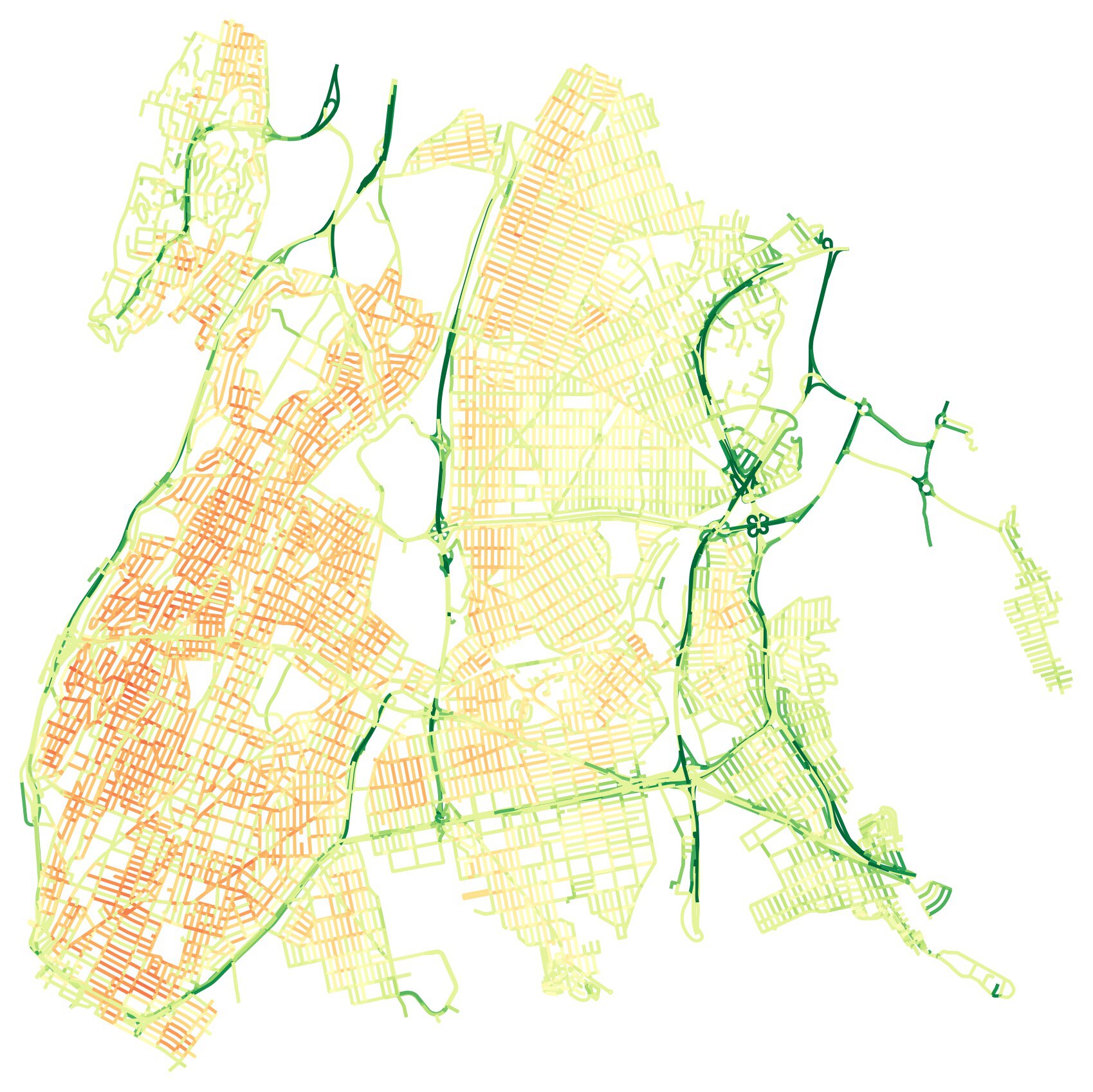}}
    \frame{\includegraphics[trim={2in, 2in, 2in, 2in},clip,width=.49\linewidth]{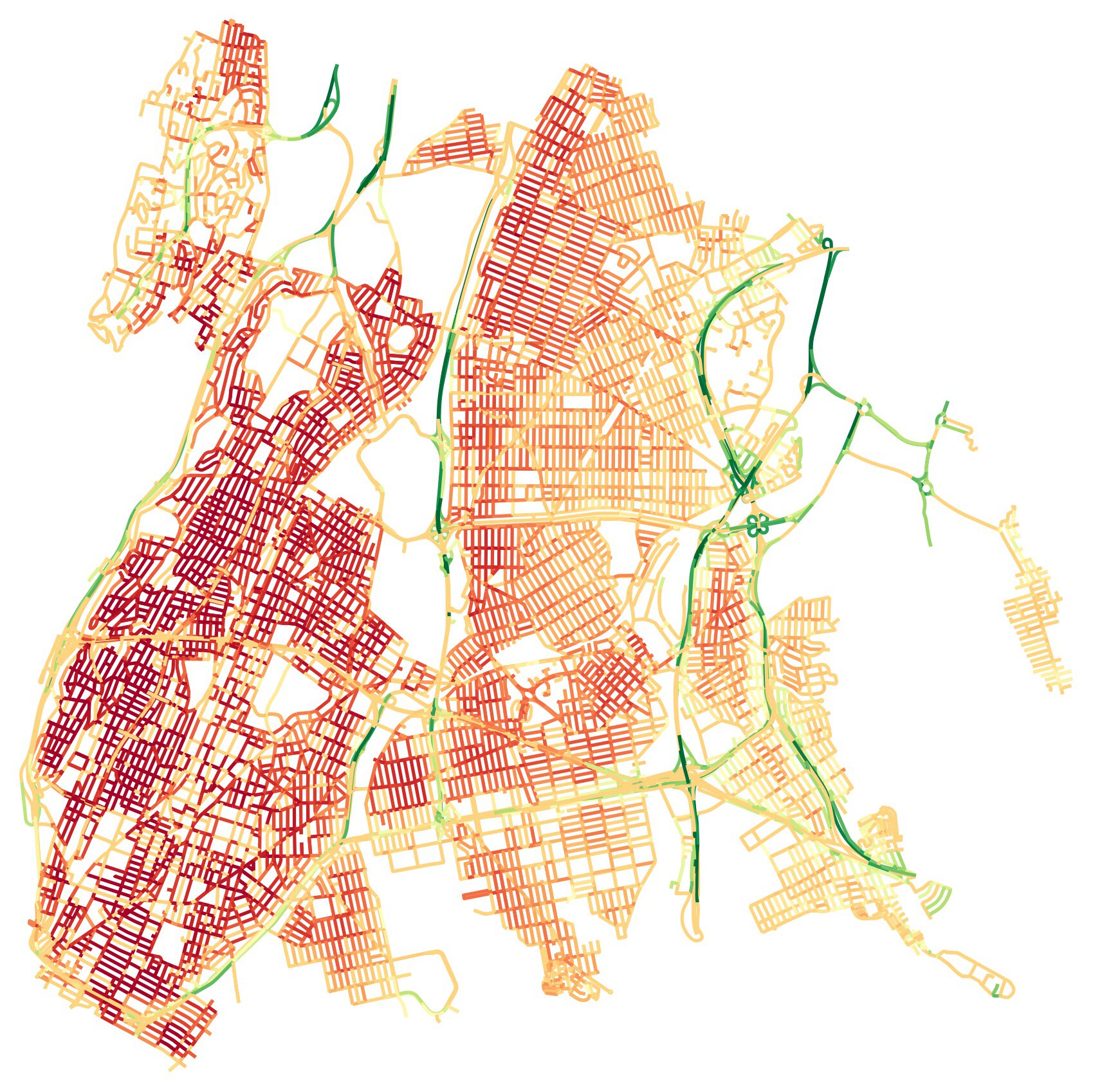}}
    
    \caption{Using our approach to dynamically model traffic flow in
    The Bronx, New York City. (left) Predicted traffic speeds for
    Monday at 4am and (right) Monday at 8am. Green (red) corresponds
    to faster (slower).}
    \label{fig:cartoon}
\end{figure}

Numerous cities are starting to equip themselves with intelligent
transportation systems, such as adaptive traffic control, that take
advantage of recent advances in computer vision and machine learning.
For example, Pittsburgh recently deployed smart traffic signals at
fifty intersections that use artificial intelligence to estimate
traffic volume and optimize traffic flow in real-time. An initial
pilot study~\cite{smith2013surtrac} indicated travel times were
reduced by 25\%, time spent waiting at signals by 40\%, number of
stops by 30\%, and emissions by 20\%. Ultimately, interest in applying
machine learning to problems in traffic management continues to grow
due to its potential for improving safety, decreasing congestion, and
reducing emissions.

\begin{figure*}
    \centering
    \frame{\includegraphics[trim=0 150px 0 80px, clip, width=.49\linewidth]{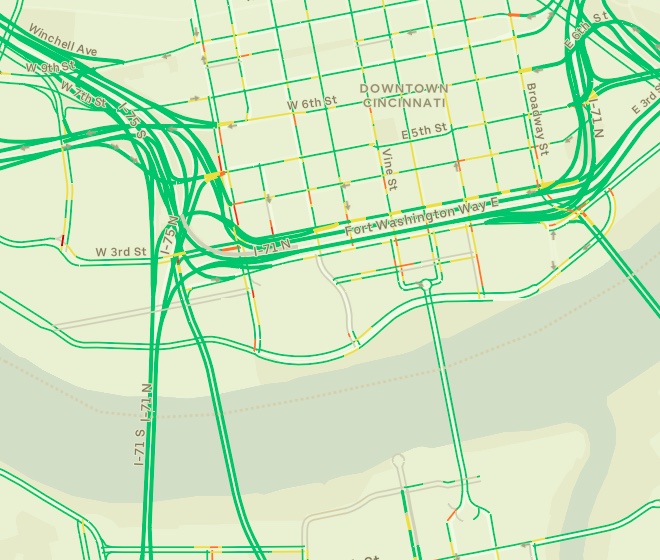}}
    \frame{\includegraphics[trim=0 150px 0 80px, clip, width=.49\linewidth]{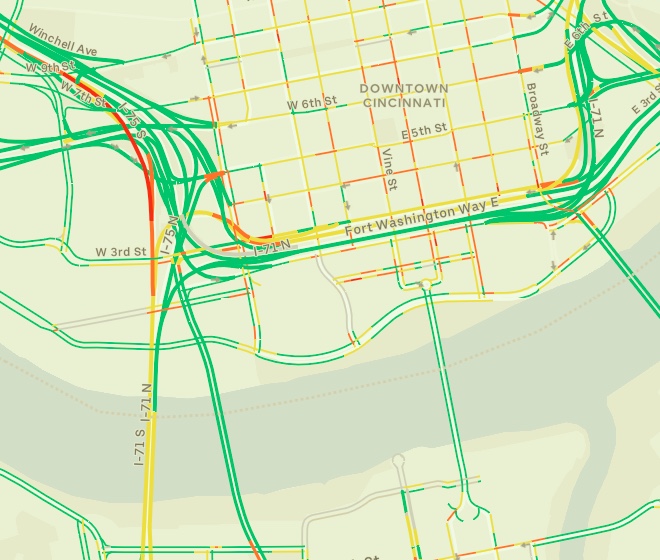}}
    
    \caption{In urban environments, traffic speeds change drastically
    throughout the day. For example, traffic congestion in downtown
    Cincinnati during the (left) early morning (12am to 7am) is
    minimal compared to (right) afternoon peak (4pm to 7pm) on a
    weekday (visualization from~\cite{uber}). Speeds are shown as
    percent from free-flow where green (red) is faster (slower).}
    \label{fig:dynamic}
\end{figure*}

Direct access to empirical traffic data is useful for planners to
analyze congestion in relation to the underlying street network, as
well as for validating models and guiding infrastructure investments.
Unfortunately, historical  information  relating  time, traffic
speeds, and street networks has typically been expensive to acquire
and limited to only primary roads. Only very recently has a large
corpus of traffic speed data been released to the public. In May of
2019, Uber Technologies, Inc. (an American multinational ridesharing
company) announced Uber Movement Speeds~\cite{uber}, a dataset of
street speeds collected from drivers of their ridesharing platform.
However, even this data has limitations, including: 1) coverage, speed
data is only available for 5 large metropolitan cities at the time of
release and 2) granularity, not all roads are traversed at all times
(or traversed at all). For example, only $29\%$ of road segments in
New York City have historical traffic data for Monday at 12pm,
considering every Monday in 2018.

In this work, our goal is to use historical traffic speeds to build a
complete model of traffic flow for a given city (\figref{dynamic}).
Traditional approaches for modeling traffic flow assume that the road
network is known (i.e., in the form of a graph reflecting the presence
and connectivity of road segments) and model road segments
individually. However, this approach is limited in that it cannot
generalize to new areas and it gives noisy estimates for road segments
with few samples. Instead, we explore how image-driven mapping can be
applied to model traffic flow directly from overhead imagery. We
envision that such a model could be used by urban planners to
understand city-scale traffic patterns when the complete road network
is unknown or insufficient empirical traffic data is available. 

We propose an automatic approach for generating dynamic maps of
traffic speeds using convolutional neural networks (CNNs). We frame
this as a multi-task learning problem, and design a network
architecture that simultaneously learns to segment roads, estimate
orientation, and predict traffic speeds. Along with overhead imagery,
the network takes as input contextual information describing the
location and time. Ultimately, the output of our method can be
considered as a local motion model that captures likely directions of
travel and corresponding travel speeds. To support training and
evaluating our methods, we introduce a new dataset that takes
advantage of a year of traffic speeds for New York City, collected
from Uber Movement Speeds.

Extensive experiments show that our approach is able to capture
complex relationships between the underlying road infrastructure and
traffic flow, enabling understanding of city-scale traffic patterns,
without requiring the road network to be known in advance. This
enables our approach to generalize to new areas. The main
contributions of this work are summarized as follows:
\begin{compactitem}
    \item introducing a new dataset for fine-grained road
      understanding, 
    \item proposing a multi-task CNN architecture for estimating a
      localized motion model directly from an overhead image,
    \item integrating location and time metadata to enable dynamic
      traffic modeling,
    \item an extensive quantitative and qualitative analysis,
      including generating dynamic city-scale travel-time maps.
\end{compactitem}
Our approach has several potential real-world applications, including
forecasting traffic speeds and providing estimates of historical
traffic speeds on roads which were not traversed during a specific
time period.

\section{Related Work}

The field of urban planning~\cite{levy2016contemporary} seeks to
understand urban environments and how they are used in order to guide
future decision making. The overarching goal is to shape the pattern
of growth as a community expands to achieve a desirable land-use
pattern. A major factor here is understanding how the environment
influences human activity. For example, research has shown how the
physical environment is associated with physical activity
(walking/cycling) and subsequently impacts
health~\cite{saelens2003environmental}.

Transportation planning is a specific subarea of urban planning that
focuses on the design of transportation systems. The goal is to
develop systems of travel that align with and promote a desired policy
of human activity~\cite{meyer1984urban}. Decisions might include how
and where to place roads, sidewalks, and other infrastructure to
minimize congestion. As such, decades of research has focused on
understanding traffic flow, i.e., the interactions of travelers and
the underlying infrastructure. For example,
Krau{\ss}~\cite{krauss1998microscopic} proposes a model of microscopic
traffic flow to understand different types of traffic congestion.
Meanwhile other work focuses on simulating urban
mobility~\cite{krajzewicz2012recent}.

In computer vision, relevant work seeks to infer properties of the
local environment directly from imagery. For example estimating
physical attributes like land cover and land
use~\cite{robinson2019large, workman2017unified}, categorizing the
type of scene~\cite{zhou2017places}, and relating appearance to
location~\cite{workman2015wide, workman2017natural,salem2020learning}.
Other work focuses on understanding urban environments. Albert et
al.~\cite{albert2017using} analyze and compare urban environments at
the scale of cities using satellite imagery. Dubey et
al.~\cite{dubey2016deep} explore the relationship between the
appearance of the physical environment and the urban perception of its
residents by predicting perceptual attributes such as \emph{safe} and
\emph{beautiful}. 

Specific to urban transportation, many studies have explored how to
identify roads and infer road networks directly from overhead
imagery~\cite{mnih2010learning, mattyus2016hd,
mattyus2017deeproadmapper, batra2019improved, sun2019leveraging}.
Recent methods in this area take advantage of convolutional neural
networks for segmenting an overhead image, then generate a graph
topology directly from the segmentation output. Mapping roads is an
important problem as it can positively impact local communities as
well as support disaster response~\cite{nourbakhsh2006mapping,
soden2014crowdsourced}. However, identifying roads is just the first
step. Other work has focused on estimating properties of roads,
including safety~\cite{song2018farsa}.

Understanding how roads are used, in particular traffic speeds, is
important for studying driver behavior, improving safety, decreasing
collisions, and aiding infrastructure planning. Therefore, several
works have tackled the problem of estimating traffic speeds from
imagery. Hua et al.~\cite{hua2018vehicle} detect, track, and estimate
traffic speeds for vehicles in traffic videos. Song et
al.~\cite{song2019remote} estimate the free-flow speed of a road
segment from a co-located overhead image and corresponding road
metadata. Van Etten~\cite{etten2020city} segments roads and estimates
road speed limits. Unlike this previous work, our goal is to
dynamically model traffic flow over time.

Similarly, traffic forecasting is an important research area. Abadi et
al.~\cite{abadi2014traffic} propose an autoregressive model for
predicting the flows of a traffic network and demonstrate the ability
to forecast near-term future traffic flows. Zhang et
al.~\cite{zhang2017deep} predict crowd flows between subregions of a
city based on historical trajectory data, weather, and events. Wang et
al.~\cite{wang2018will} propose a deep learning framework for
path-based travel time estimation. These methods typically assume
prior knowledge of the spatial connectivity of the road network,
unlike our work which operates directly on overhead imagery.

\section{A Large Traffic Speeds Dataset}
\label{sec:dataset}

\begin{figure}
    \centering
    \frame{\includegraphics[width=1\linewidth]{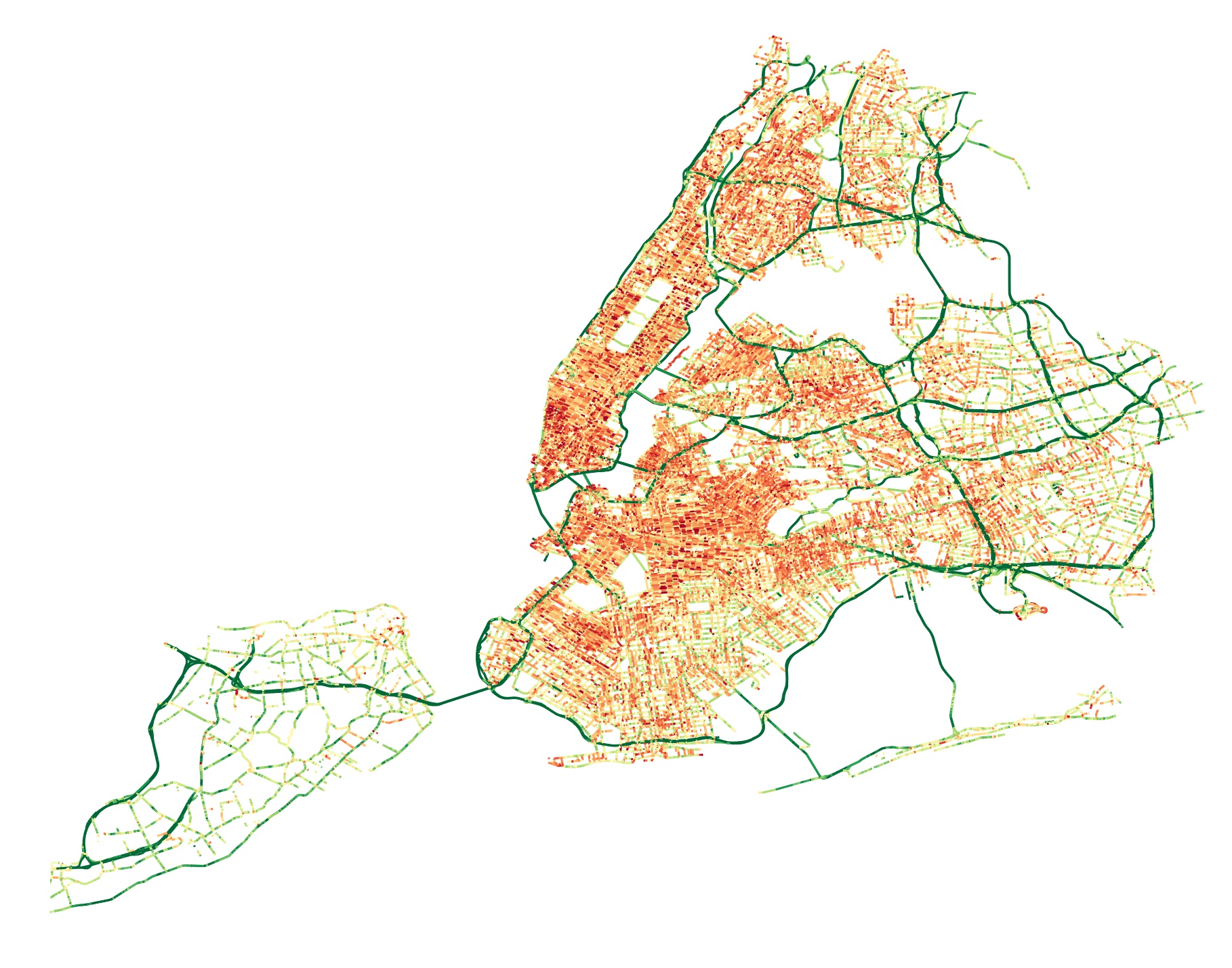}}
    \caption{Example traffic speed data from Uber Movement Speeds for
    New York City (visualized as free-flow speeds for January 2018).}
    \label{fig:movement}
\end{figure}

To support training and evaluating our methods, we introduce the
Dynamic Traffic Speeds (DTS) dataset that takes advantage of a year of
historical traffic speeds for New York City. Our traffic speed data is
collected from Uber Movement Speeds~\cite{uber}, a dataset of publicly
available aggregated speed data over road segments at hourly
frequencies.

\subsection{Uber Movement Speeds}

During rideshare trips, Uber (via their Uber Driver application)
frequently collects GPS data including latitude, longitude, speed,
direction, and time. While this data supports many functionalities, it
is also stored for offline processing, where it is aggregated and used
to derive speed data. Additionally, Uber uses OpenStreetMap as the
source of their underlying map data (i.e., road
network).\footnote{\url{ https://www.openstreetmap.org}}

Given the map and GPS data as input, an extensive process is used to
1) match the GPS data to locations on the street network, 2) compute
segment traversal speeds using the matched data, and 3) aggregate
speeds along each segment. Please refer to the
whitepaper~\cite{movement} for a detailed overview of this process.
Ultimately, the publicly released data includes the road segment
identifier and average speed along that segment at an hourly
resolution. Note that bidirectional roads, represented as line
strings, are twinned and a speed estimate is provided for each
direction.

\subsection{Augmenting with Overhead Imagery}

To support our methods, we generated an aligned dataset of overhead
images, contained road segments, and historical traffic speeds. We
started by collecting road geometries and speed data from Uber
Movement Speeds for New York City during the 2018 calendar year. This
resulted in over 292 million records, or approximately 22GB (not
including road geometries). For reference, there are around 290
thousand road segments in NYC when considering bidirectional roads.
\figref{movement} shows the free-flow speed along these segments for
January 2018, where free-flow speed is defined as the 85th percentile
of all recorded traffic speeds.

Starting from a bounding box around New York City, we generated a set
of non-overlapping tiles using the standard XYZ style spherical
Mercator tile. For each tile, we identified the contained road
segments, extracted the corresponding speed data along those segments,
and downloaded an overhead image from Bing Maps (filtering out tiles
that do not contain any roads). This process resulted in approximately
\num{12000} $\num{1024} \times \num{1024}$ overhead images at
$\sim0.3$ meters / pixel. We partitioned these non-overlapping tiles
into 85\% training, 5\% validation, and 10\% testing. The result is a
large dataset containing overhead images (over 12 billion pixels),
road geometries, and traffic speed data (along with other road
attributes).

\begin{figure}
  \centering
  \setlength\tabcolsep{1pt}
    \begin{tabular}{ccc}
      \includegraphics[width=.32\linewidth]{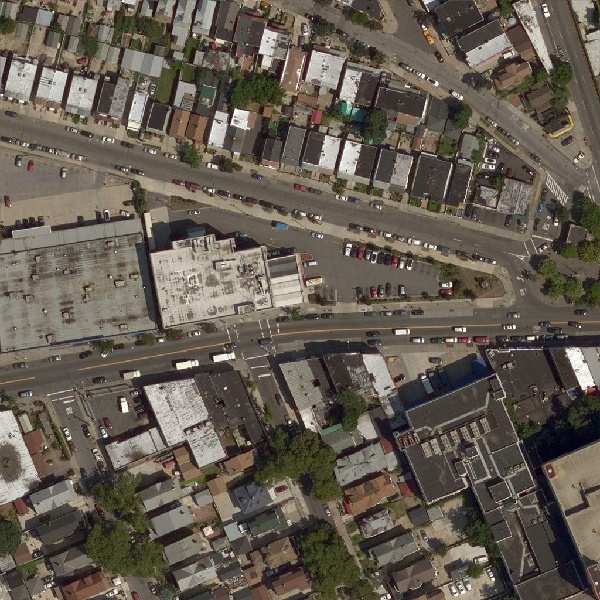} &
      \includegraphics[width=.32\linewidth]{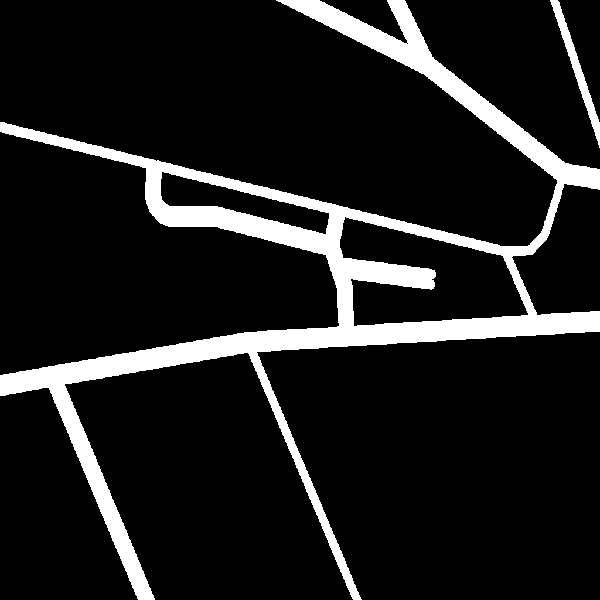} &
      \includegraphics[width=.32\linewidth]{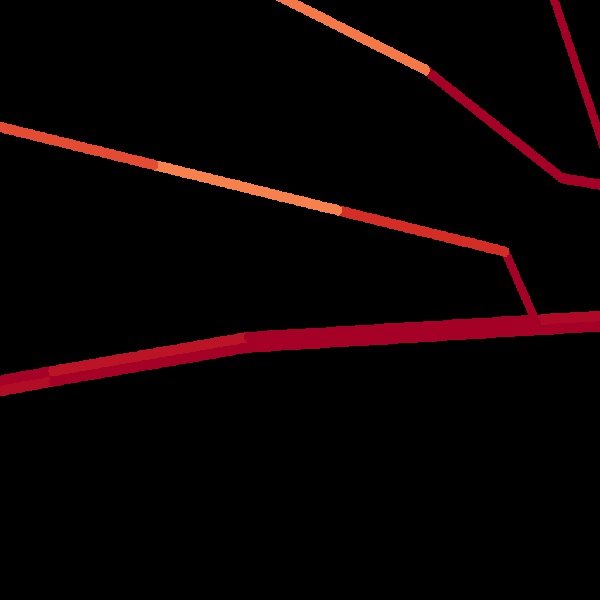} \\
      
      \includegraphics[width=.32\linewidth]{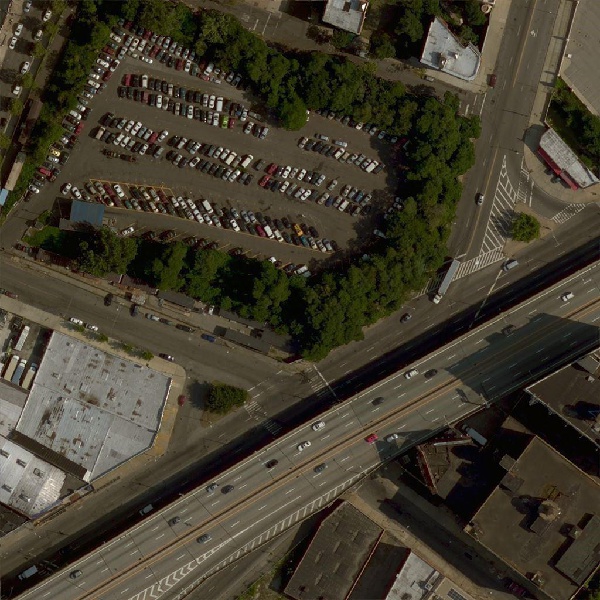} &
      \includegraphics[width=.32\linewidth]{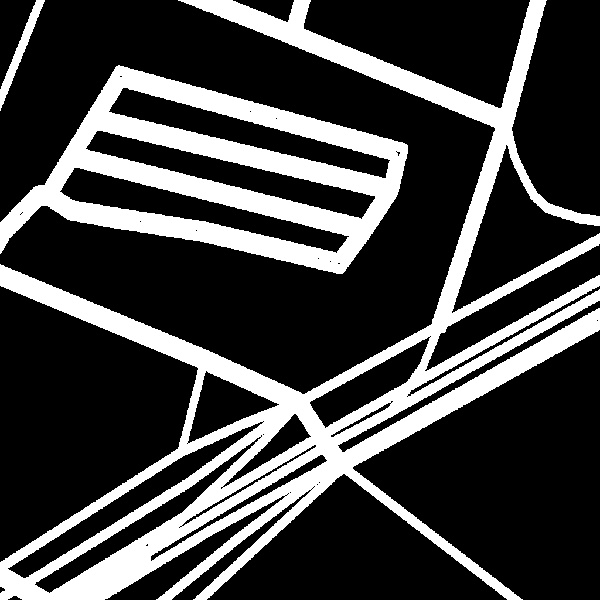} &
      \includegraphics[width=.32\linewidth]{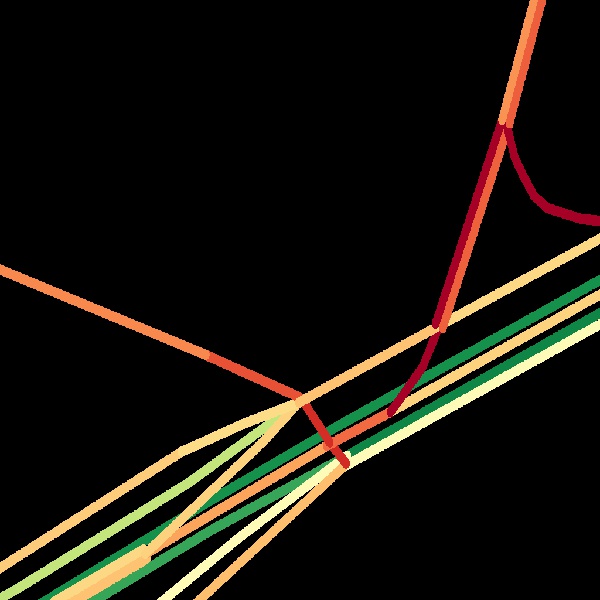} \\

    \end{tabular}

    \caption{Our dataset: (left) example image, (middle) road mask,
    and (right) speed mask (rendered using a random time). Notice that
    speed data is not available for every road at every time.}

  \label{fig:data}
\end{figure}

\figref{data} shows some example data: from left to right, an overhead
image, the corresponding road mask, and a mask characterizing the
traffic speeds at a given time. Notice that, depending on the time,
not all roads have valid speed data. For this visualization, road
geometries are buffered (converted to polygons) with two meter half
width.

\subsection{Aggregating Traffic Speeds}

For a single road segment, there are a possible \num{8760} ($365
\times 24$) unique recorded speeds for that segment over the course of
a year. When considering all roads, this is a large amount of data.
For this work, we instead aggregate speed data for each road segment
using day of week and hour of day, retaining the number of samples
observed (i.e., the number of days per year that traffic was recorded
at that time on a particular segment). This reduces the number of
possible traffic speeds to \num{168} ($7 \times 24$) per segment.

\subsection{Discussion}

While the current version of the dataset includes only New York City,
we are actively working towards expanding it to include other cities
where traffic speed data is available (e.g., London, Cincinnati).
Further, we plan to incorporate other contextual road attributes
(e.g., type of road, surface material, number of lanes) so that our
dataset is useful for other tasks in fine-grained road understanding.
Our hope is that this dataset will inspire further work in computer
vision directed towards traffic modeling, with a positive impact on
urban planning and minimizing traffic congestion.

\begin{figure*}
    \centering
    \includegraphics[width=1\linewidth]{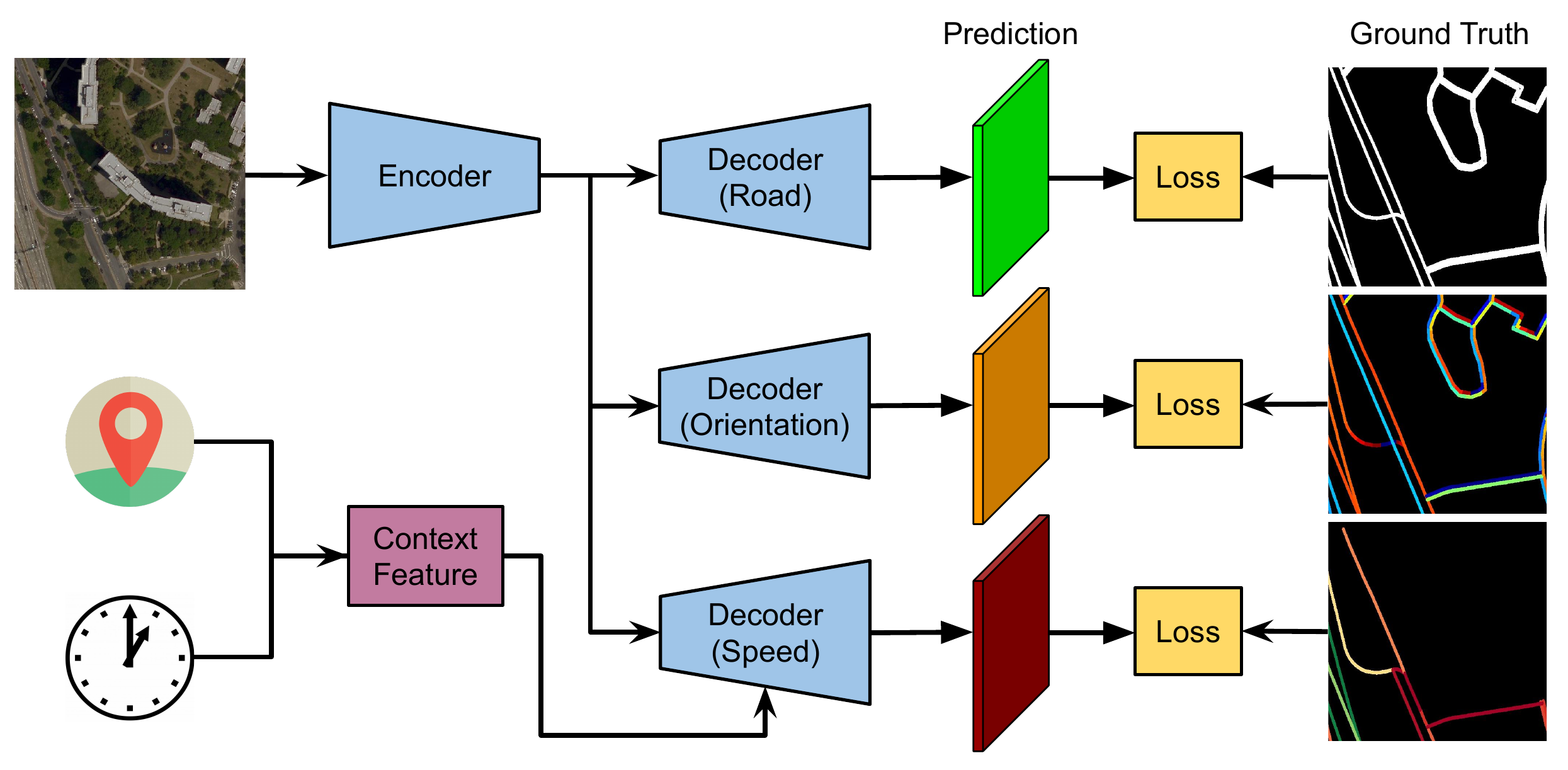}
    \caption{An overview of our network architecture.}
    \label{fig:architecture}
\end{figure*}

\section{Modeling Traffic Flow}
\label{sec:method}

We propose a novel CNN that fuses high-resolution overhead imagery,
location, and time to estimate a dynamic model of traffic flow. We can
think of our task as learning a conditional probability distribution
over velocity, $P(\vec{v}| S(l), l, t)$, where $l$ is a
latitude-longitude coordinate, $S(l)$ is an overhead image centered at
that location, and $t$ represents the time.

\subsection{Architecture Overview}

We propose a multi-task architecture that simultaneously solves three
pixel-wise labeling tasks: segmenting roads, estimating orientation,
and predicting traffic speeds. Our network (\figref{architecture}) has
three inputs: a location, $l$, the time, $t$, and an overhead image,
$S(l)$, centered at $l$ (of size $H \times W \times C$). We build on a
modern, lightweight, semantic segmentation architecture,
LinkNet~\cite{chaurasia2017linknet}, that follows an encoder/decoder
approach with skip connections between every layer. Specifically, we
use LinkNet-34, which is LinkNet with a ResNet-34~\cite{he2016deep}
encoder. For our purposes, we modify the baseline architecture to a
multi-task version, with a shared encoder and separate decoder for
each task. Though we use LinkNet, our approach would work with any
modern encoder/decoder segmentation architecture.

\paragraph{Integrating Location and Time Context} 

To make location and time-dependent traffic flow predictions, we
integrate location and time into the final speed classification
layers. We represent location as normalized latitude/longitude
coordinates ($\mu=0, \sigma^2=1$). Time is parameterized as day of
week (0-6) and hour of day (0-23). Each time dimension is represented
as an embedding lookup with an embedding dimension of 3. To form a
context feature, we concatenate the parameterized location and time
embedding outputs together. The context feature is then tiled and
fused in the decoder at each of the final two convolutional layers
(concatenated on as additional feature channels).

\paragraph{Loss Function} 

We simultaneously optimize the entire network, in an end-to-end
manner, for all three tasks. The final loss function becomes:
\begin{equation}
    \mathcal{L} = \mathcal{L}_{road} + \mathcal{L}_{orientation} + \mathcal{L}_{speed} + \alpha_{r}\mathcal{L}_{reg},
\end{equation}
where $\mathcal{L}_{road}$, $\mathcal{L}_{orientation}$, and
$\mathcal{L}_{speed}$ correspond to the task-specific objective
function for road segmentation, orientation estimation, and traffic
speed prediction, respectively. Additionally, $\mathcal{L}_{reg}$ is a
regularization term that is weighted by a scalar $\alpha_r$. In the
following sections we detail the specifics of each task, including the
architecture and respective loss terms.

\subsection{Identifying Presence of Roads}

The objective of the first decoder is to segment roads in an overhead
image. We represent this as a binary classification task (road vs.\
not road) resulting in a single output per pixel ($H \times W \times
1$).  The output is passed through a \emph{sigmoid} activation
function. We follow recent trends in state-of-the-art road
segmentation~\cite{zhou2018d} and formulate the objective function as
a combination of multiple individual elements. The objective is:
\begin{equation}
    \mathcal{L}_{road} = \mathcal{L}_{bce} + (1 -\mathcal{L}_{dice}),
\end{equation}
where $\mathcal{L}_{bce}$ is binary cross entropy, a standard loss
function used in binary classification tasks, and
$\mathcal{L}_{dice}$ is the dice coefficient, which measures spatial
overlap.

\subsection{Estimating Direction of Travel}

The objective of the second decoder is to estimate the direction of
travel along the road at each pixel. We represent this as a
multi-class classification task over $K$ angular bins, resulting in
$K$ outputs per pixel ($H \times W \times K$). A \emph{softmax}
activation function is applied to the output. For this task, the
per-pixel loss function is categorical cross entropy:
\begin{equation}
    \mathcal{L}_{orientation} = - \log(G(S(l); \Theta)(y)),
\end{equation}
where $G(S(l); \Theta)$ represents our CNN as a function that outputs
a probability distribution over the $K$ angular bins and $y$ indicates
the true label. We compute road orientation, $\theta$, as the angle
the road direction vector makes with the positive X axis. The valid
range of values is between $-\pi$ and $\pi$ and we generate angular
bins by dividing this space uniformly.

\subsection{Predicting Traffic Speeds}

The objective of the final decoder is to estimate local traffic
speeds, taking into account the imagery, the location, and the time.
Instead of predicting a single speed value for each pixel, we make
angle-dependent predictions. The road speed decoder has $K$ outputs
per pixel ($H \times W \times K$) and a \emph{softplus} output
activation, $\log(1 + \exp(x))$, to ensure positivity. For a given
road angle $\theta$, we compute the estimated speed as an
orientation-weighted average using $w_\mu = e^{k \cos(\theta - \mu)}$
as the weight for each bin where $\mu$ is the angle of the
corresponding bin and $k=25$ is a fixed smoothing factor. Weights are
normalized to sum to one. Note that we can predict angle-dependent
speeds using either the true angle if known, or the predicted angle. 

For traffic speed estimation, we minimize the Charbonnier loss (also
known as the Pseudo-Huber loss):
\begin{equation}
    \mathcal{L}_{speed} = \delta^2 (\sqrt{1 + (a/\delta)^2} - 1),
\end{equation}
where $y$ and $\hat{y}$ are the observed and predicted values,
respectively, and $a = y - \hat{y}$ is their residual. The Charbonnier
loss is a smooth approximation of the Huber loss, where $\delta$
controls the steepness. In addition, we add a regularization term,
$\mathcal{L}_{reg}$, to reduce noise and encourage spatial smoothness.
For this we use the anisotropic version of total variation, $f(x) =
(x_{i+1,j} - x_{i,j})^2 + (x_{i,j+1} - x_{i,j})^2$, averaged over all
pixels, $i,j$, in the raw output. 

\paragraph{Region Aggregation} The target labels for traffic speed are
provided as averages over road segments, which means we cannot use a
traditional per-pixel loss. The na\"ive approach would be to assume
the speed is constant across the entire road segment, which would lead
to over-smoothing and incorrect predictions. Instead, we use a variant
of the region aggregation layer~\cite{jacobs2018weakly}, adapted to
compute the average of the per-pixel estimated speeds over the
segment. We optimize our network to generate per-pixel speeds such
that the segment averages match the true speed. In practice we predict
angle-dependent speeds, compute the orientation-weighted average, and
then apply region aggregation; finally, computing the average loss
over road segments.

\subsection{Implementation Details}

Our methods are implemented using PyTorch~\cite{pytorch} and optimized
using RAdam~\cite{liu2020variance} ($\lambda = 10^{-3}$) with
Lookahead~\cite{zhang2019lookahead} ($k=5, \alpha=0.5$). We initialize
the encoder with weights from a network pretrained on ImageNet. For
fairness, we train all networks with a batch size of 6 for 50 epochs,
on random crops of size $640 \times 640$. We use $K=16$ angular bins
and set $\alpha_{r} = 10^{-2}$, chosen empirically. For this work, we
set $\delta=2$. Our networks are trained in a dynamic manner; instead
of rendering a segmentation mask at every possible time, for every
training image, we sample a time during training and dynamically
render the speed mask. The alternative would be to pregenerate over a
million segmentation masks. Additionally, we train the orientation and
speed estimation decoders in a sparse manner by sampling pixels along
road segments (every one meter along the segment and up to two meters
on each side, perpendicularly) and computing orientation from
corresponding direction vectors. For road segmentation, we buffer the
road geometries (two meter half width) and do not sample.  Model
selection is performed using the validation set. 

\section{Evaluation}

We train and evaluate our method using the dataset described in
\secref{dataset}. We primarily evaluate our model for traffic speed
estimation, but present quantitative and qualitative results for both
road segmentation and orientation estimation.

\subsection{Ablation Study}

We conducted an extensive ablation study to evaluate the impact of
various components of our proposed architecture. For evaluation, we
use the reserved test set, but evaluate on a single timestep per image
(randomly selected from the observed traffic speeds using a fixed
seed). When computing metrics, we represent speeds using kilometers
per hour and average predictions along each road segment before
comparing to the ground truth.

\subsubsection{Impact of Multi-Task Learning} 

For our first experiment, we quantify the impact of multi-task
learning for estimating traffic speeds. In other words, we evaluate
whether or not simultaneously performing the road segmentation and
orientation estimation tasks improves the results for estimating
traffic speeds. We compare our full method (\secref{method}) to
variants with subsets of the multi-task components. The results of
this experiment are shown in \tblref{multi} for three metrics:
root-mean-square error (RMSE), mean absolute error (MAE) and the
coefficient of determination ($R^2$). 

As observed, the multi-task nature of our architecture improves the
final speed predictions. Adding the road segmentation and orientation
estimation tasks improves results over a baseline that only estimates
traffic speed, with the best performing model integrating both tasks.
Our results are in line with previous work~\cite{zhang2014facial} that
demonstrates multi-task learning can be helpful when the auxiliary
tasks are related to the primary task. For the remainder of the
ablation study, we only consider our full architecture that performs
all three tasks.

\begin{table}[t]
  \centering
  \caption{Evaluating the impact of multi-task learning for traffic speed estimation.}
  \begin{tabular}{@{}cccrrr@{}}
    \toprule
    Road & Orientation & RMSE & MAE & $R^2$ \\
    \bottomrule
    \ding{55} & \ding{55} & 10.87 & 8.35 & 0.442 \\
    \ding{55} & \ding{51} & 10.78 & 8.21 & 0.452 \\
    \ding{51} & \ding{55} & 10.73 & 8.19 & 0.456 \\
    \ding{51} & \ding{51} & \textbf{10.66} & \textbf{8.10} & \textbf{0.464} \\
    \bottomrule
  \end{tabular}
  \label{tbl:multi}
\end{table}

\subsubsection{Impact of Region Aggregation}

Next, we consider how region aggregation affects traffic speed
estimation. As described in \secref{method}, the target speeds for the
traffic speed estimation task are averages over each road segment.
Here we compare two approaches for training using these labels: 1)
na\"ively replicating the target labels spatially across the entire
road segment, and 2) our approach that integrates a variant of the
region aggregation layer~\cite{jacobs2018weakly}, which enables
prediction of per-pixel speeds such that the segment averages match
the ground-truth speed label for that segment. 

\begin{figure}
    \centering
    \begin{subfigure}[t]{.32\linewidth}
        \centering
        \includegraphics[trim={2.9in .2in .2in 2.9in},clip,width=1\linewidth]{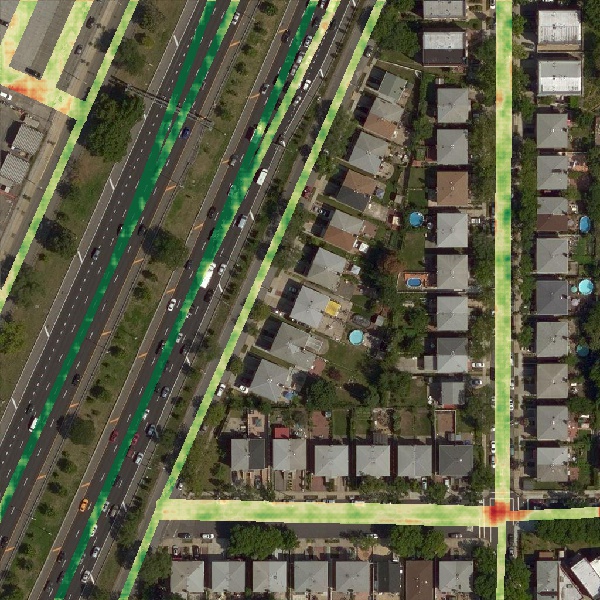}
        \includegraphics[trim={2.9in .2in .2in 2.9in},clip,width=1\linewidth]{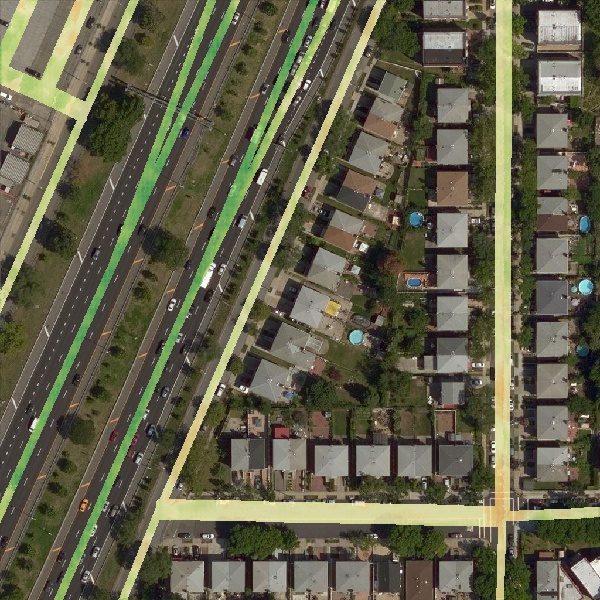}
    \end{subfigure}
    \begin{subfigure}[t]{.32\linewidth}
        \centering
        \includegraphics[trim={2.4in 1.2in .2in 1.4in},clip,width=1\linewidth]{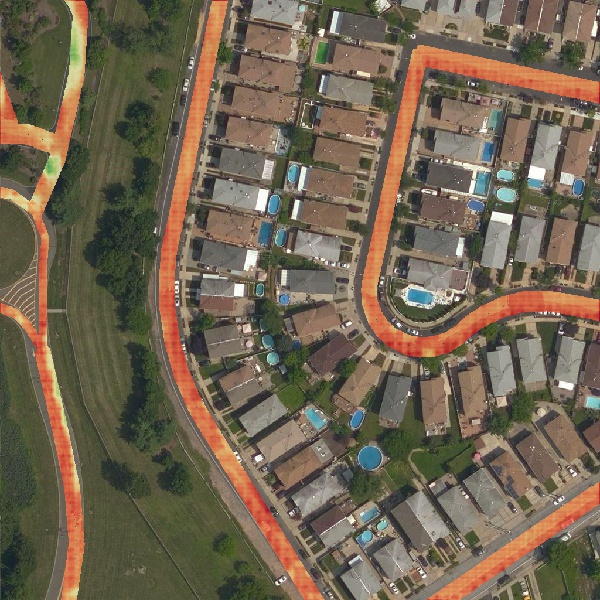}
        \includegraphics[trim={2.4in 1.2in .2in 1.4in},clip,width=1\linewidth]{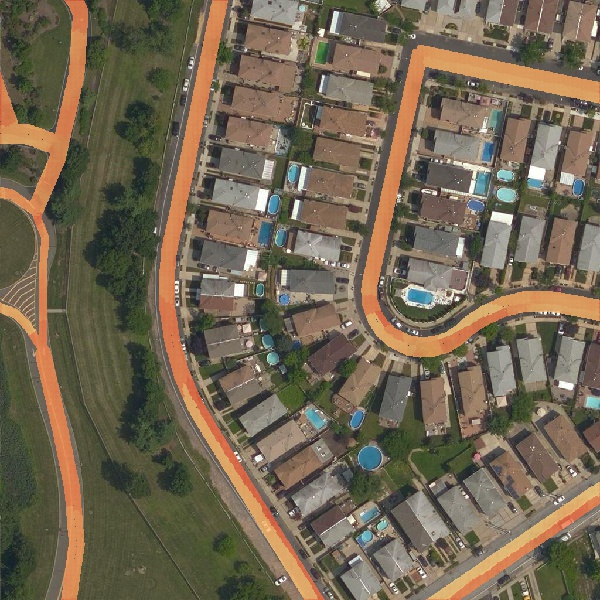}
    \end{subfigure}
    \begin{subfigure}[t]{.32\linewidth}
        \centering
        \includegraphics[trim={2.2in .4in .4in 2.2in},clip,width=1\linewidth]{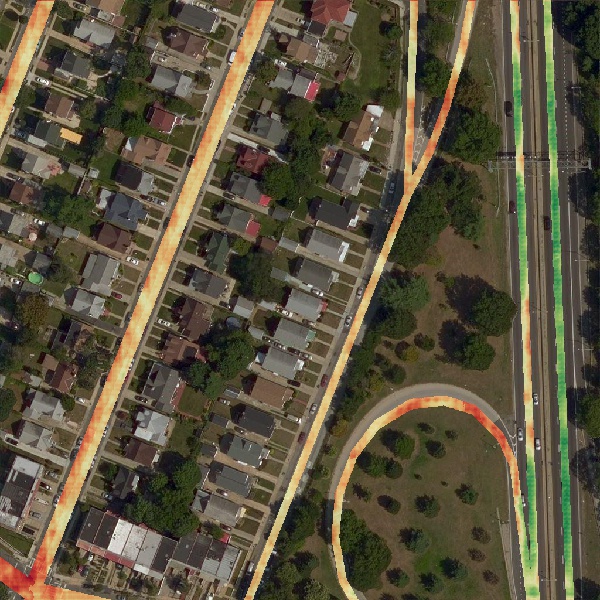}
        \includegraphics[trim={2.2in .4in .4in 2.2in},clip,width=1\linewidth]{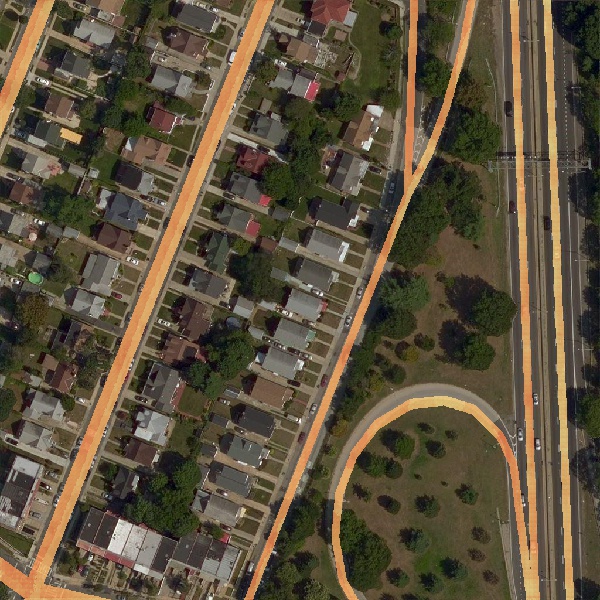}
    \end{subfigure}

    \caption{Qualitative examples showing the impact of region
    aggregation (top) versus no aggregation (bottom).}
    \label{fig:disaggregation}
\end{figure}

To evaluate both approaches, we average predictions along each road
segment and compare to the true traffic speed label. The baseline
method achieves an RMSE score of $11.10$, which is worse than our
method (RMSE = $10.66$). Additionally, we show some qualitative
results of the two approaches in \figref{disaggregation}. Our method
which incorporates region aggregation (top) is better able to capture
real-world properties of traffic speed, such as slowing down at
intersections or around corners. For the remainder of the ablation
study, we only consider methods which were optimized and evaluated
using region aggregation.

\subsubsection{Impact of Location and Time Context}

Finally, we evaluate how integrating location and time context impacts
our traffic speed predictions. For this experiment, we compare against
several baseline methods that share many low-level components with our
proposed architecture. Our full model includes all three components
\emph{image}, \emph{loc}, and \emph{time}. For the metadata only
approaches, those without \emph{image},  we use our proposed
architecture, but omit all layers prior to concatenating in the
context feature. 

The results of this experiment are shown in \tblref{ablation}. Both
location and time improve the resulting traffic speed predictions. Our
method, which integrates overhead imagery, location, and time,
outperforms all other models. Additionally, we show results for road
segmentation (F1 score) and orientation estimation (top-1 accuracy).
These tasks do not rely on location and time, so their performance is
comparable, but our method still performs best.

\begin{table}
  \centering
  \caption{Evaluating the impact of location and time context.}
  \begin{tabular}{@{}lccc@{}}
    \toprule
    & \multicolumn{1}{c}{Road} & \multicolumn{1}{c}{Orientation} & \multicolumn{1}{c}{Speed} \\
    & \multicolumn{1}{c}{(F1 Score)} & \multicolumn{1}{c}{(Accuracy)} & \multicolumn{1}{c}{(RMSE)} \\
    \bottomrule
    {\em loc}               & \multicolumn{1}{c}{--} & \multicolumn{1}{c}{--} &  13.38 \\
    {\em time}              & \multicolumn{1}{c}{--} & \multicolumn{1}{c}{--} &  14.06 \\
    {\em loc, time}         & \multicolumn{1}{c}{--} & \multicolumn{1}{c}{--} &  13.14 \\
    \midrule
    {\em image}             & 0.796 & 75.05\% & 11.35 \\
    {\em image, loc}        & 0.798 & 75.63\% & 10.95 \\
    {\em image, time}       & 0.798 & 76.04\% & 10.68 \\
    {\em image, loc, time}  & \textbf{0.800} & \textbf{76.32\%} & \textbf{10.66} \\
    \bottomrule
  \end{tabular}
  \label{tbl:ablation}
\end{table}

\begin{figure}
    \centering
    \begin{subfigure}{0.45\linewidth}
        \includegraphics[width=1\linewidth]{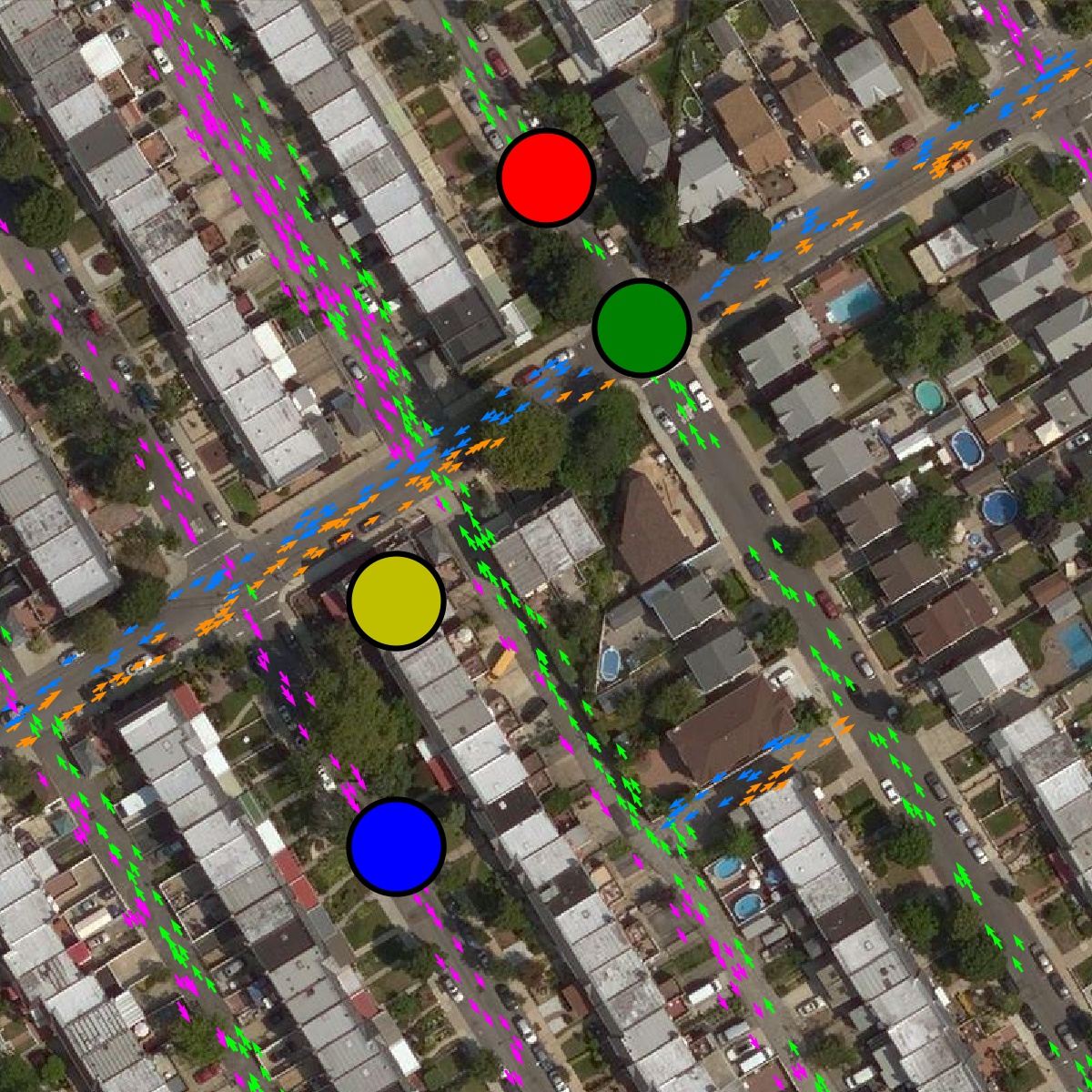}
    \end{subfigure}
    \begin{subfigure}{0.54\linewidth}
        \includegraphics[width=.49\linewidth]{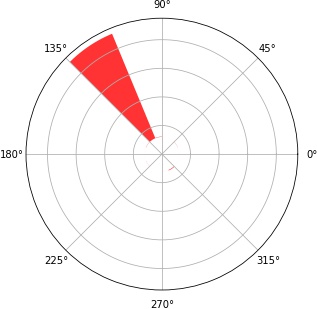}
        \includegraphics[width=.49\linewidth]{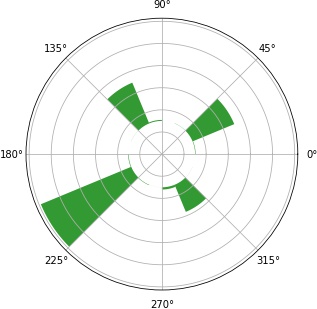}
        \includegraphics[width=.49\linewidth]{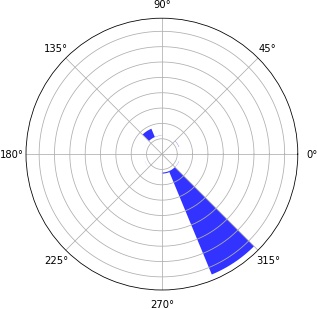}
        \includegraphics[width=.49\linewidth]{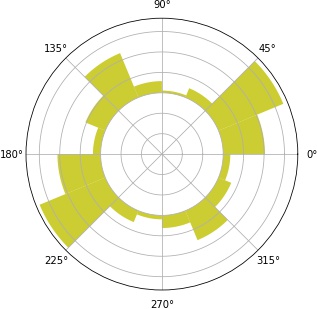}
    \end{subfigure}
    \caption{Estimating directions of travel. (left) An overhead image
    and estimated orientation represented as a flow field. (right)
    Predicted distributions over orientation for corresponding dots in
    the image. (top, right) The predicted distribution for the green
    dot correctly identifies multiple possible directions of travel.
    This makes sense as the location in the image is at an
    intersection. }
    \label{fig:orientation}
\end{figure}

\begin{figure*}
  \centering
    \begin{subfigure}{.34\linewidth}
        \includegraphics[width=.49\linewidth]{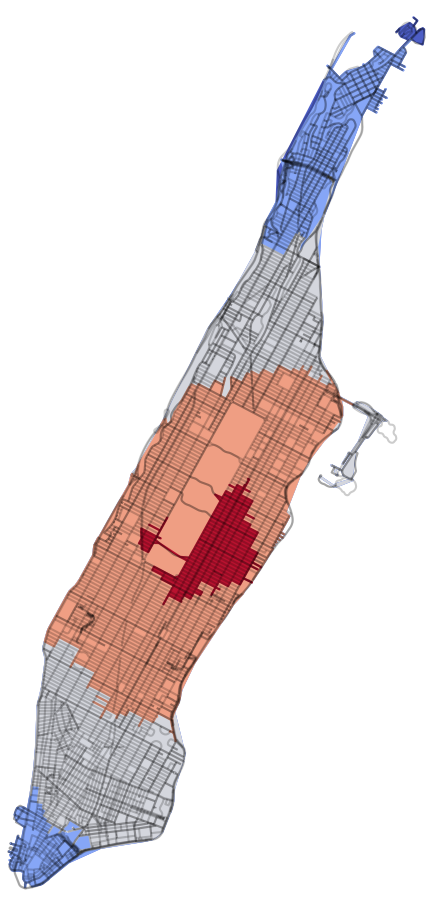}
        \includegraphics[width=.49\linewidth]{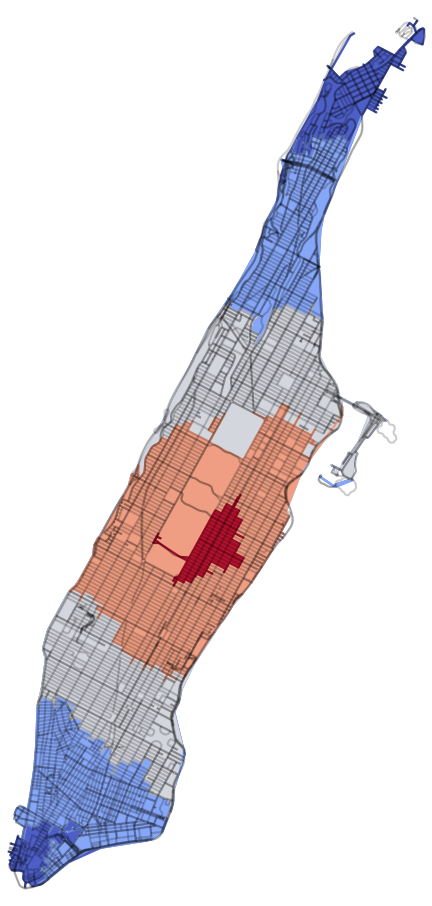}
        \caption{Manhattan}
    \end{subfigure}
    \begin{subfigure}{.617\linewidth}
        \includegraphics[width=.49\linewidth]{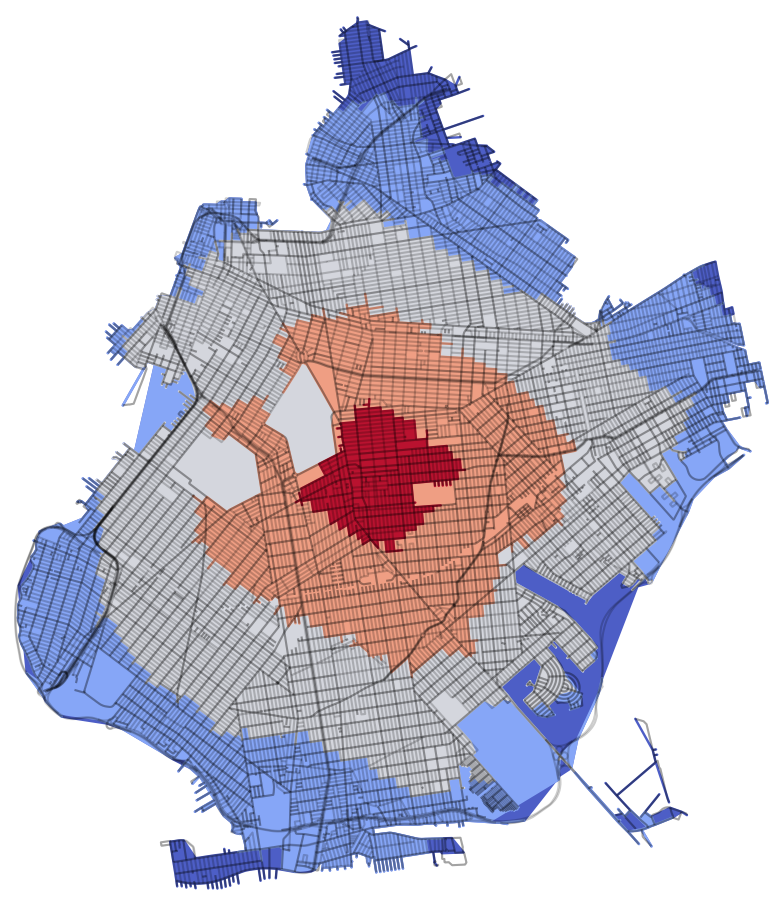}
        \includegraphics[width=.49\linewidth]{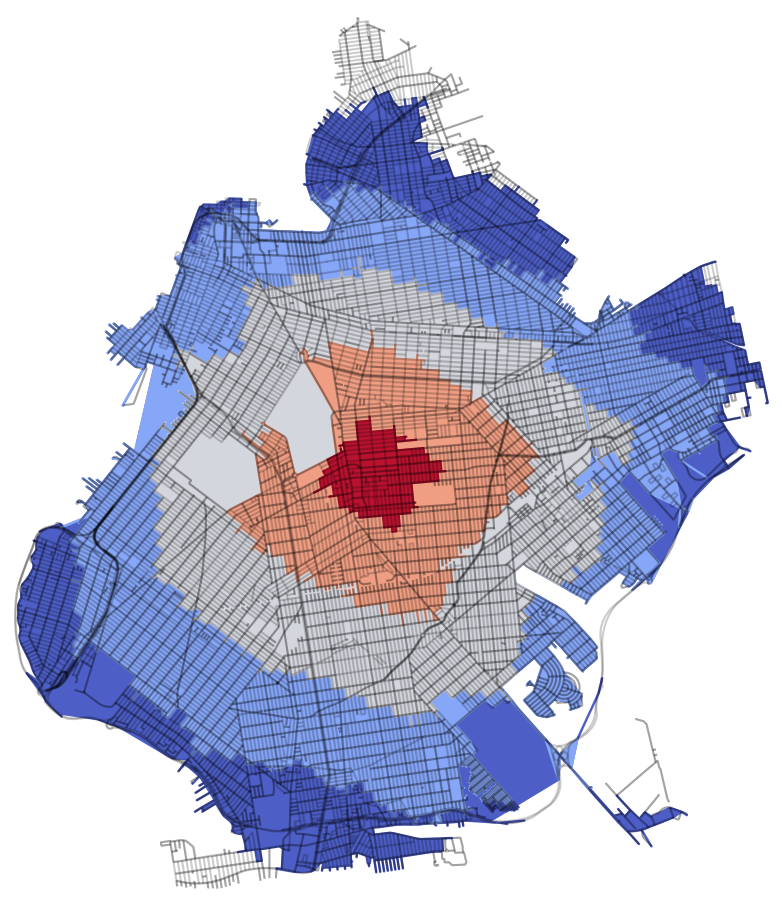}
        \caption{Brooklyn}
    \end{subfigure}
    
    \caption{Isochrone maps obtained by using our approach.
    Isocontours represent the amount of time that it would take to
    travel to all locations within the borough, starting from the
    center. For each borough the maps correspond to (left) Monday at
    4am and (right) Monday at 8am. As anticipated, travel is more
    restrictive at 8am, likely corresponding to rush hour traffic.}

  \label{fig:isochrone}
\end{figure*}

\subsection{Visualizing Traffic Flow}

In this section, we qualitatively evaluate our proposed methods
ability to capture spatial and temporal patterns in traffic flow.
First, we examine how well our approach is able to estimate directions
of travel. \figref{orientation} (left) visualizes the predicted
per-pixel orientation for an overhead image, overlaid as a vector
field (colored by predicted angle). As observed, our method is able to
capture primary directions of travel, including differences in one way
and bidirectional roads. Additionally, \figref{orientation} (right)
shows radial histograms representing the predicted distributions over
orientation for corresponding color-coded dots in the overhead image.
For example, the predicted distribution for the green dot (top,
right), which represents an intersection in the image, correctly
identifies several possible directions of travel. Alternatively, the
predicted distribution for the yellow dot is more uniform, which makes
sense as the location is not a road. 

\begin{figure}
    \centering
    \newcommand{\centered}[1]{\begin{tabular}{l} #1 \end{tabular}}
    \setlength\tabcolsep{1.5pt}
    \begin{tabular}{cc}
         \centered{\includegraphics[width=.345\linewidth]{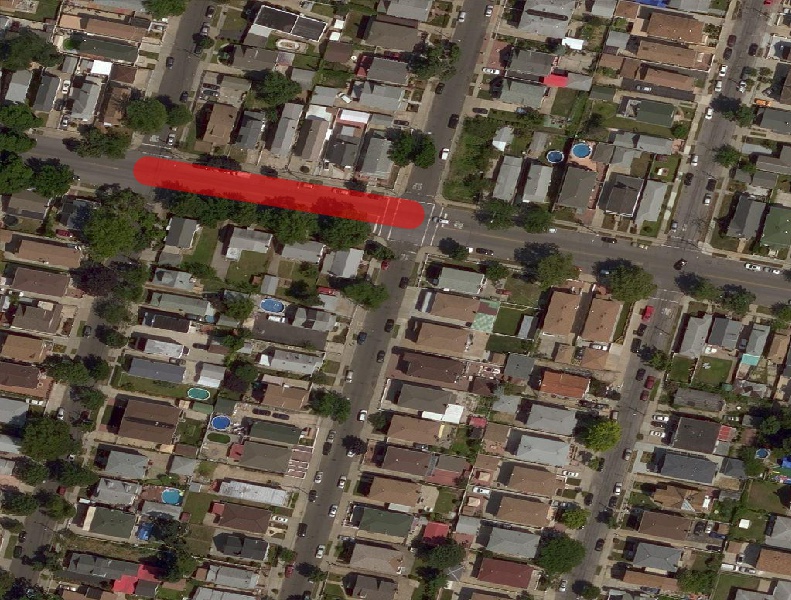}} &
         \centered{\includegraphics[width=.645\linewidth]{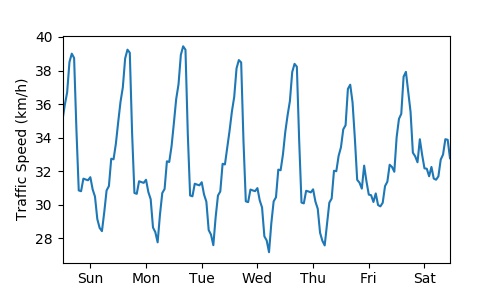}} \\ 
    \end{tabular}
    \caption{Visualizing how predicted speeds for a segment capture
    daily and weekly trends.}
    \label{fig:segment}
\end{figure}

Next, we examine how our model captures temporal trends in traffic
flow. \figref{segment} visualizes traffic speed predictions for a
residential road segment colored in red. \figref{segment} (right)
shows the predicted speeds for this segment versus the day of the
week, with each day representing twenty four hours. As observed, the
predicted speeds capture temporal trends both daily, and over the
course of the full week. Finally, \figref{cartoon} shows predicted
traffic speeds for The Bronx, New York City for Monday at 4am (left)
versus Monday at 8am (right). As expected, there is a large slow down
likely corresponding to rush hour. These results demonstrate that our
model is capturing meaningful patterns in both space and time.

\subsection{Application: Generating Travel Time Maps}

Our approach can be used to generate dynamic traffic maps at the scale
of a city. To demonstrate this, we generate travel time maps at
different times. We use OSMnx~\cite{boeing2017osmnx}, a library for
modeling street networks from OpenStreetMap, to represent the
underlying street network topology for New York City as a graph. Our
approach is as follows. For each image in our dataset, we estimate
traffic speeds at a given time. Then we update the edge weights of the
graph (corresponding to each road segment) to represent travel times
using the length of each segment in meters and our traffic speed
predictions. For any road segment not represented in our dataset, we
use the average predicted traffic speed for that time.
\figref{isochrone} shows several results, visualized as isochrone maps
that depict areas of equal travel time.

\section{Conclusion}

Understanding traffic flow is important and has many potential
implications. We developed a method for dynamically modeling traffic
flow using overhead imagery. Though our method incorporates time, a
unique overhead image is not required for every timestamp. Our model
is conditioned on location and time metadata and can be used to render
dynamic city-scale traffic maps. To support our efforts, we introduced
a novel dataset for fine-grained road understanding. Our hope is that
this dataset will inspire further work in the area of image-driven
traffic modeling.  By simultaneously optimizing for road segmentation,
orientation estimation, and traffic speed prediction, our method can
be applied for understanding traffic flow patterns in novel areas.
Potential applications of our method include assisting urban planners,
augmenting routing engines, and for providing a general understanding
of how to traverse an environment. \\

\ifcvprfinal

\noindent\textbf{Acknowledgements} We gratefully acknowledge the
financial support of NSF CAREER grant IIS-1553116.

\fi

{\small
\bibliographystyle{ieee_fullname}
\bibliography{biblio}

\begin{thebibliography}{10}\itemsep=-1pt

\bibitem{uber}
{Data retrieved from Uber Movement, (c) 2020 Uber Technologies, Inc.}
\newblock \url{https://movement.uber.com}.

\bibitem{movement}
{Uber Movement}: Speeds calculation methodology.
\newblock Technical report, Uber Technologies, Inc., 2019.

\bibitem{abadi2014traffic}
Afshin Abadi, Tooraj Rajabioun, and Petros~A Ioannou.
\newblock Traffic flow prediction for road transportation networks with limited
  traffic data.
\newblock {\em IEEE Transactions on Intelligent Transportation Systems},
  16(2):653--662, 2014.

\bibitem{albert2017using}
Adrian Albert, Jasleen Kaur, and Marta~C Gonzalez.
\newblock Using convolutional networks and satellite imagery to identify
  patterns in urban environments at a large scale.
\newblock In {\em ACM SIGKDD International Conference on Knowledge Discovery
  and Data Mining}, 2017.

\bibitem{batra2019improved}
Anil Batra, Suriya Singh, Guan Pang, Saikat Basu, CV Jawahar, and Manohar
  Paluri.
\newblock Improved road connectivity by joint learning of orientation and
  segmentation.
\newblock In {\em IEEE Conference on Computer Vision and Pattern Recognition},
  2019.

\bibitem{boeing2017osmnx}
Geoff Boeing.
\newblock {OSM}nx: New methods for acquiring, constructing, analyzing, and
  visualizing complex street networks.
\newblock {\em Computers, Environment and Urban Systems}, 65:126--139, 2017.

\bibitem{chaurasia2017linknet}
Abhishek Chaurasia and Eugenio Culurciello.
\newblock {LinkNet}: Exploiting encoder representations for efficient semantic
  segmentation.
\newblock In {\em IEEE Visual Communications and Image Processing}, 2017.

\bibitem{dubey2016deep}
Abhimanyu Dubey, Nikhil Naik, Devi Parikh, Ramesh Raskar, and C{\'e}sar~A
  Hidalgo.
\newblock Deep learning the city: Quantifying urban perception at a global
  scale.
\newblock In {\em European Conference on Computer Vision}, 2016.

\bibitem{etten2020city}
Adam~Van Etten.
\newblock City-scale road extraction from satellite imagery v2: Road speeds and
  travel times.
\newblock In {\em IEEE Winter Conference on Applications of Computer Vision},
  2020.

\bibitem{he2016deep}
Kaiming He, Xiangyu Zhang, Shaoqing Ren, and Jian Sun.
\newblock Deep residual learning for image recognition.
\newblock In {\em IEEE Conference on Computer Vision and Pattern Recognition},
  2016.

\bibitem{hua2018vehicle}
Shuai Hua, Manika Kapoor, and David~C Anastasiu.
\newblock Vehicle tracking and speed estimation from traffic videos.
\newblock In {\em CVPR Workshop on AI City Challenge}, 2018.

\bibitem{jacobs2018weakly}
Nathan Jacobs, Adam Kraft, Muhammad~Usman Rafique, and Ranti~Dev Sharma.
\newblock A weakly supervised approach for estimating spatial density functions
  from high-resolution satellite imagery.
\newblock In {\em ACM SIGSPATIAL International Conference on Advances in
  Geographic Information Systems}, 2018.

\bibitem{krajzewicz2012recent}
Daniel Krajzewicz, Jakob Erdmann, Michael Behrisch, and Laura Bieker.
\newblock Recent development and applications of sumo-simulation of urban
  mobility.
\newblock {\em International Journal On Advances in Systems and Measurements},
  5(3\&4), 2012.

\bibitem{krauss1998microscopic}
Stefan Krau{\ss}.
\newblock {\em Microscopic modeling of traffic flow: Investigation of collision
  free vehicle dynamics}.
\newblock PhD thesis, University of Cologne, 1998.

\bibitem{levy2016contemporary}
John~M Levy.
\newblock {\em Contemporary urban planning}.
\newblock Taylor \& Francis, 2016.

\bibitem{liu2020variance}
Liyuan Liu, Haoming Jiang, Pengcheng He, Weizhu Chen, Xiaodong Liu, Jianfeng
  Gao, and Jiawei Han.
\newblock On the variance of the adaptive learning rate and beyond.
\newblock In {\em International Conference on Learning Representations}, 2020.

\bibitem{mattyus2017deeproadmapper}
Gell{\'e}rt M{\'a}ttyus, Wenjie Luo, and Raquel Urtasun.
\newblock {DeepRoadMapper}: Extracting road topology from aerial images.
\newblock In {\em IEEE International Conference on Computer Vision}, 2017.

\bibitem{mattyus2016hd}
Gell{\'e}rt M{\'a}ttyus, Shenlong Wang, Sanja Fidler, and Raquel Urtasun.
\newblock {HD Maps}: Fine-grained road segmentation by parsing ground and
  aerial images.
\newblock In {\em IEEE Conference on Computer Vision and Pattern Recognition},
  2016.

\bibitem{meyer1984urban}
Michael~D Meyer and Eric~J Miller.
\newblock {\em Urban transportation planning: a decision-oriented approach}.
\newblock McGraw-Hill, 1984.

\bibitem{mnih2010learning}
Volodymyr Mnih and Geoffrey~E Hinton.
\newblock Learning to detect roads in high-resolution aerial images.
\newblock In {\em European Conference on Computer Vision}, 2010.

\bibitem{nourbakhsh2006mapping}
Illah Nourbakhsh, Randy Sargent, Anne Wright, Kathryn Cramer, Brian McClendon,
  and Michael Jones.
\newblock Mapping disaster zones.
\newblock {\em Nature}, 439(7078):787, 2006.

\bibitem{pytorch}
Adam Paszke, Sam Gross, Francisco Massa, Adam Lerer, James Bradbury, Gregory
  Chanan, Trevor Killeen, Zeming Lin, Natalia Gimelshein, Luca Antiga, et~al.
\newblock {PyTorch}: An imperative style, high-performance deep learning
  library.
\newblock In {\em Advances in Neural Information Processing Systems}, 2019.

\bibitem{robinson2019large}
Caleb Robinson, Le Hou, Kolya Malkin, Rachel Soobitsky, Jacob Czawlytko, Bistra
  Dilkina, and Nebojsa Jojic.
\newblock Large scale high-resolution land cover mapping with multi-resolution
  data.
\newblock In {\em IEEE Conference on Computer Vision and Pattern Recognition},
  2019.

\bibitem{saelens2003environmental}
Brian~E Saelens, James~F Sallis, and Lawrence~D Frank.
\newblock Environmental correlates of walking and cycling: findings from the
  transportation, urban design, and planning literatures.
\newblock {\em Annals of Behavioral Medicine}, 25(2):80--91, 2003.

\bibitem{salem2020learning}
Tawfiq Salem, Scott Workman, and Nathan Jacobs.
\newblock Learning a dynamic map of visual appearance.
\newblock In {\em IEEE Conference on Computer Vision and Pattern Recognition},
  2020.

\bibitem{schrank2019urban}
David Schrank, Bill Eisele, and Tim Lomax.
\newblock 2019 urban mobility report.
\newblock Technical report, Texas A\&M Transportation Institute, 2019.

\bibitem{smith2013surtrac}
Stephen~F Smith, Gregory Barlow, Xiao-Feng Xie, and Zachary~B Rubinstein.
\newblock {SURTRAC}: Scalable urban traffic control.
\newblock {\em Transportation Research Board Annual Meeting}, 2013.

\bibitem{soden2014crowdsourced}
Robert Soden and Leysia Palen.
\newblock From crowdsourced mapping to community mapping: The post-earthquake
  work of {OpenStreetMap} {H}aiti.
\newblock In {\em International Conference on the Design of Cooperative
  Systems}, 2014.

\bibitem{song2019remote}
Weilian Song, Tawfiq Salem, Hunter Blanton, and Nathan Jacobs.
\newblock Remote estimation of free-flow speeds.
\newblock In {\em IEEE International Geoscience and Remote Sensing Symposium},
  2019.

\bibitem{song2018farsa}
Weilian Song, Scott Workman, Armin Hadzic, Xu Zhang, Eric Green, Mei Chen,
  Reginald Souleyrette, and Nathan Jacobs.
\newblock {FARSA}: Fully automated roadway safety assessment.
\newblock In {\em IEEE Winter Conference on Applications of Computer Vision},
  2018.

\bibitem{sprung2018transportation}
Michael~J Sprung, Sonya Smith-Pickel, et~al.
\newblock Transportation statistics annual report.
\newblock Technical report, Bureau of Transportation Statistics, 2018.

\bibitem{sun2019leveraging}
Tao Sun, Zonglin Di, Pengyu Che, Chun Liu, and Yin Wang.
\newblock Leveraging crowdsourced {GPS} data for road extraction from aerial
  imagery.
\newblock In {\em IEEE Conference on Computer Vision and Pattern Recognition},
  2019.

\bibitem{triplett2016american}
Tim Triplett, Rob Santos, Sandra Rosenbloom, and Brian Tefft.
\newblock American driving survey: 2014--2015.
\newblock Technical report, The American Automobile Association, 2016.

\bibitem{wang2018will}
Dong Wang, Junbo Zhang, Wei Cao, Jian Li, and Yu Zheng.
\newblock When will you arrive? {E}stimating travel time based on deep neural
  networks.
\newblock In {\em AAAI Conference on Artificial Intelligence}, 2018.

\bibitem{workman2015wide}
Scott Workman, Richard Souvenir, and Nathan Jacobs.
\newblock Wide-area image geolocalization with aerial reference imagery.
\newblock In {\em IEEE International Conference on Computer Vision}, 2015.

\bibitem{workman2017natural}
Scott Workman, Richard Souvenir, and Nathan Jacobs.
\newblock Understanding and mapping natural beauty.
\newblock In {\em IEEE International Conference on Computer Vision}, 2017.

\bibitem{workman2017unified}
Scott Workman, Menghua Zhai, David~J. Crandall, and Nathan Jacobs.
\newblock A unified model for near and remote sensing.
\newblock In {\em IEEE International Conference on Computer Vision}, 2017.

\bibitem{zhang2017deep}
Junbo Zhang, Yu Zheng, and Dekang Qi.
\newblock Deep spatio-temporal residual networks for citywide crowd flows
  prediction.
\newblock In {\em AAAI Conference on Artificial Intelligence}, 2017.

\bibitem{zhang2019lookahead}
Michael~R Zhang, James Lucas, Geoffrey Hinton, and Jimmy Ba.
\newblock Lookahead optimizer: k steps forward, 1 step back.
\newblock In {\em Advances in Neural Information Processing Systems}, 2019.

\bibitem{zhang2014facial}
Zhanpeng Zhang, Ping Luo, Chen~Change Loy, and Xiaoou Tang.
\newblock Facial landmark detection by deep multi-task learning.
\newblock In {\em European Conference on Computer Vision}, 2014.

\bibitem{zhou2017places}
Bolei Zhou, Agata Lapedriza, Aditya Khosla, Aude Oliva, and Antonio Torralba.
\newblock Places: A 10 million image database for scene recognition.
\newblock {\em IEEE Transactions on Pattern Analysis and Machine Intelligence},
  40(6):1452--1464, 2017.

\bibitem{zhou2018d}
Lichen Zhou, Chuang Zhang, and Ming Wu.
\newblock {D-LinkNet}: {LinkNet} with pretrained encoder and dilated
  convolution for high resolution satellite imagery road extraction.
\newblock In {\em CVPR Workshop on DeepGlobe}, 2018.

\end{thebibliography}
}

\newpage
\null
\vskip .375in
\twocolumn[{%
  \begin{center}
    \textbf{\Large Supplemental Material : \\ Dynamic Traffic Modeling from Overhead Imagery}
  \end{center}
  \vspace*{24pt}
}]
\setcounter{section}{0}
\setcounter{equation}{0}
\setcounter{figure}{0}
\setcounter{table}{0}
\makeatletter
\renewcommand{\theequation}{S\arabic{equation}}
\renewcommand{\thefigure}{S\arabic{figure}}
\renewcommand{\thetable}{S\arabic{table}}

This document contains additional details and experiments related to
our methods.

\section{Dynamic Traffic Speeds Dataset}

\begin{figure}
    \centering
    \includegraphics[width=1\linewidth]{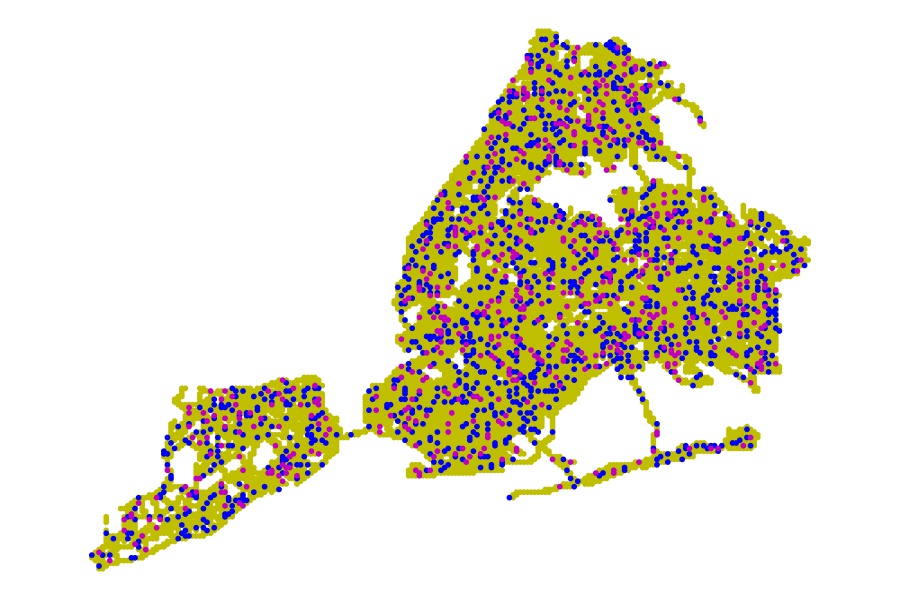}
    \caption{Coverage of our dataset. Each dot corresponds to a
    non-overlapping overhead image (yellow training, blue testing,
    magenta validation).}
    \label{fig:coverage}
\end{figure}

\begin{figure*}[t]
  \centering
  \setlength\tabcolsep{1pt}
    \begin{tabular}{ccccc}
      \includegraphics[width=.196\linewidth]{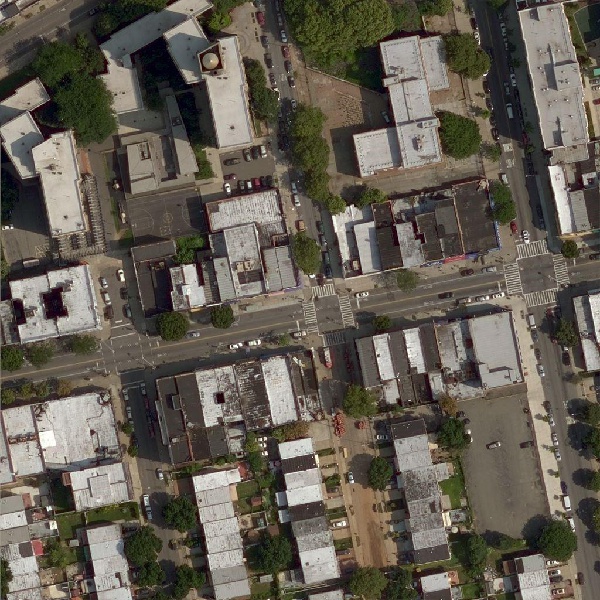} &
      \includegraphics[width=.196\linewidth]{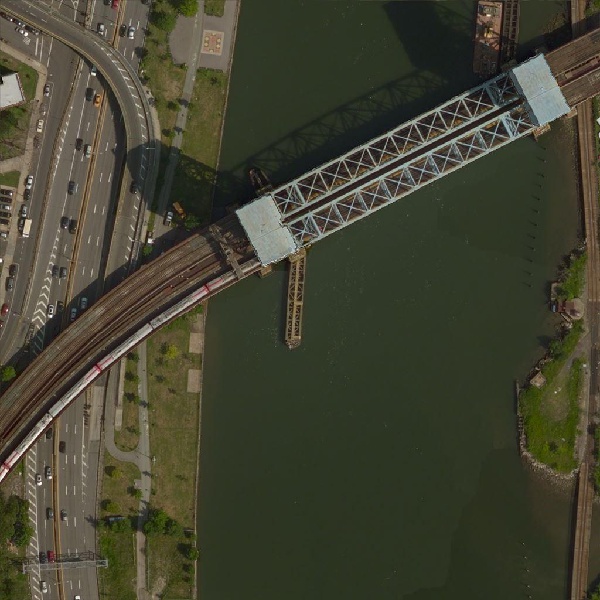} &
      \includegraphics[width=.196\linewidth]{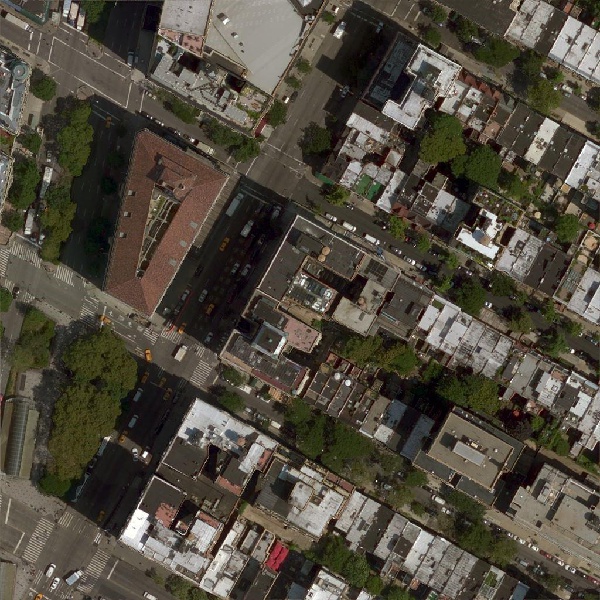} &
      \includegraphics[width=.196\linewidth]{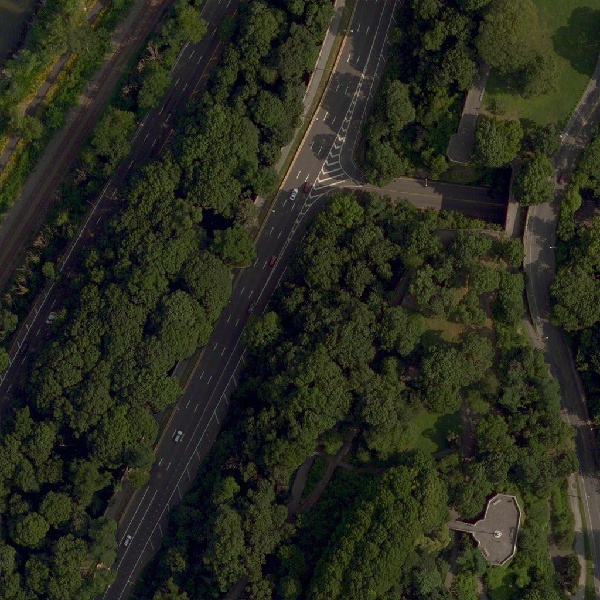} &
      \includegraphics[width=.196\linewidth]{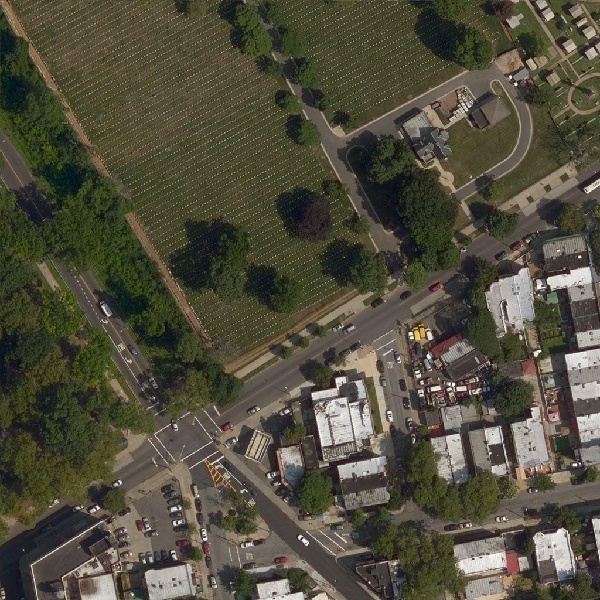} \\
      
      \includegraphics[width=.196\linewidth]{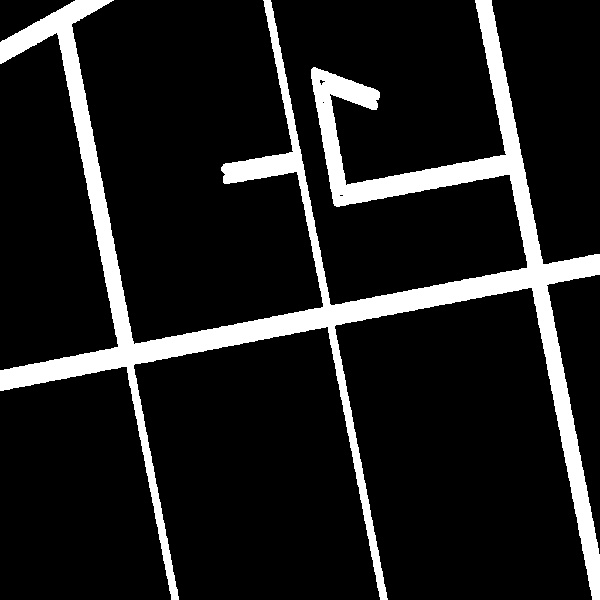} &
      \includegraphics[width=.196\linewidth]{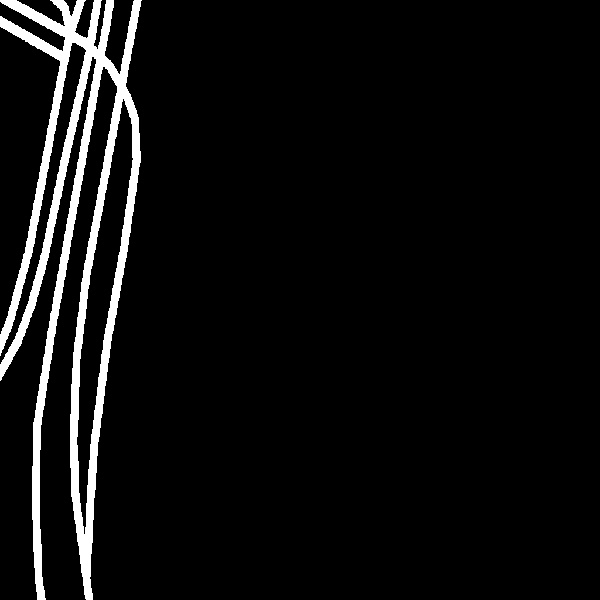} &
      \includegraphics[width=.196\linewidth]{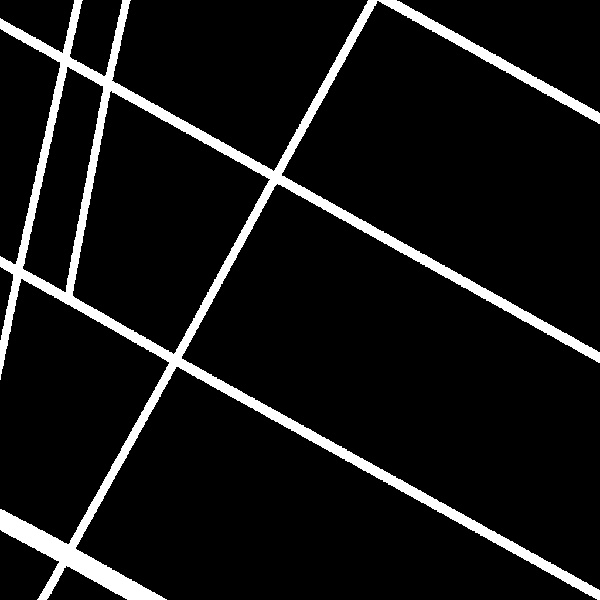} &
      \includegraphics[width=.196\linewidth]{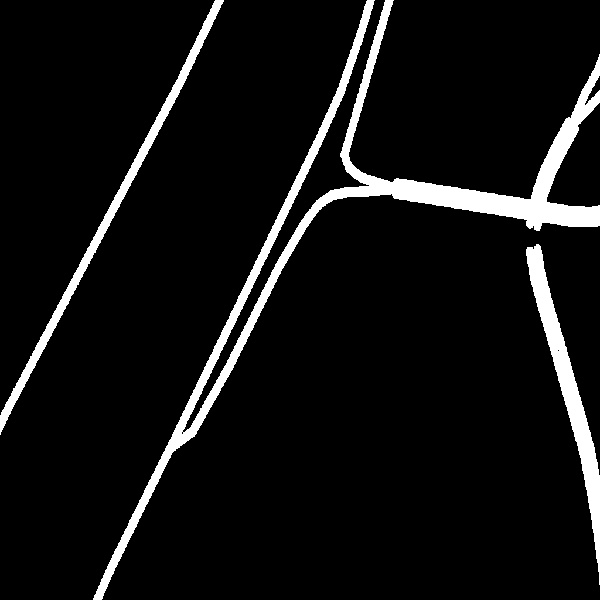} &
      \includegraphics[width=.196\linewidth]{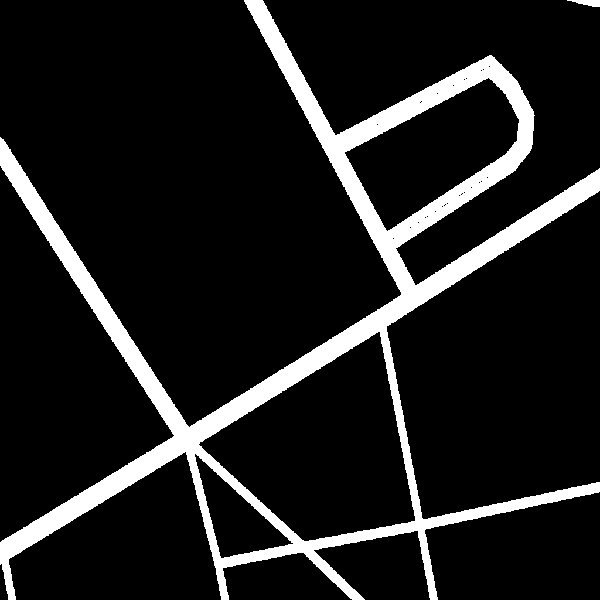} \\
      
      \includegraphics[width=.196\linewidth]{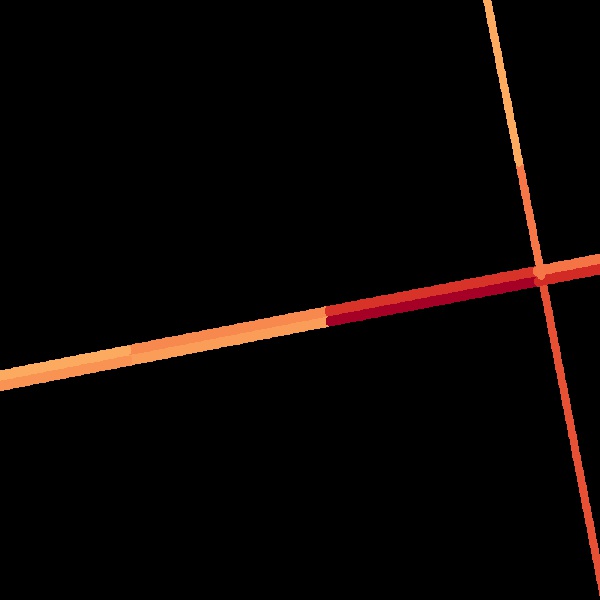} &
      \includegraphics[width=.196\linewidth]{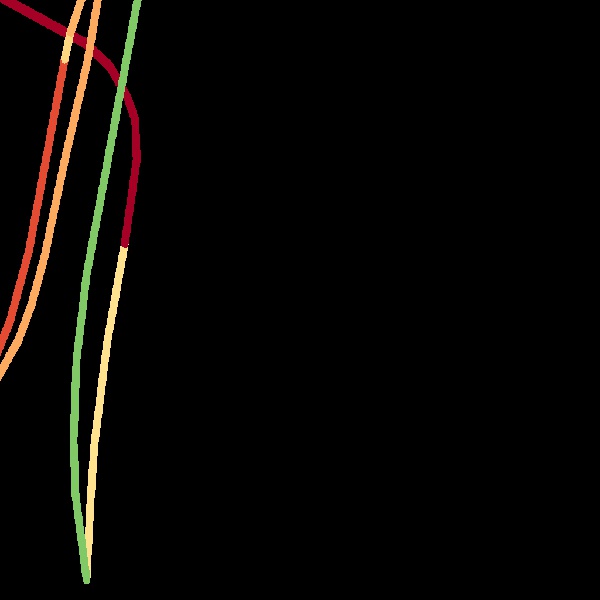} &
      \includegraphics[width=.196\linewidth]{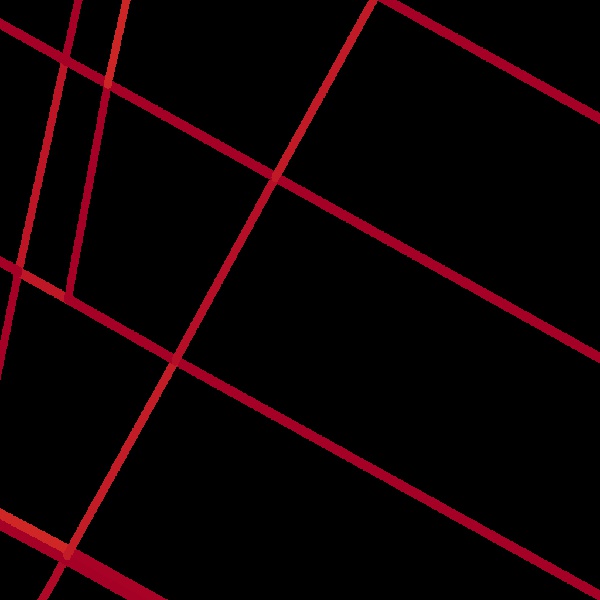} &
      \includegraphics[width=.196\linewidth]{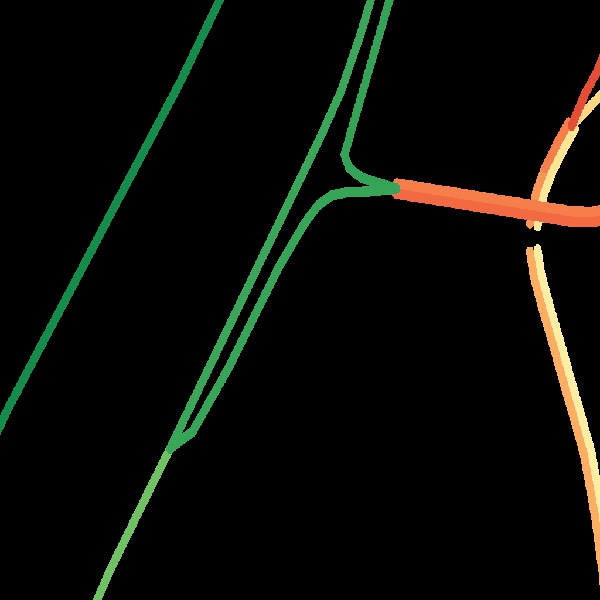} &
      \includegraphics[width=.196\linewidth]{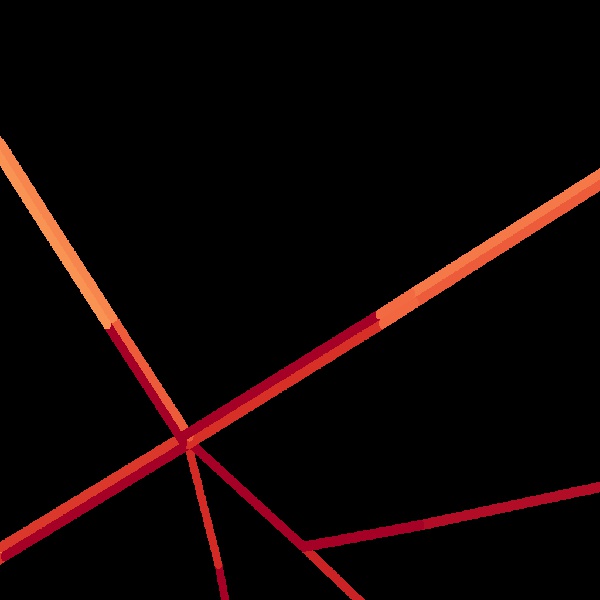} \\
    \end{tabular}

    \caption{Examples from our dataset: (top) image, (middle) road
    mask, and (bottom) speed mask rendered using a random time (where
    red is slower and green faster). Notice that historical speed data
    is not available for every road at every time.}

  \label{fig:data_supp}
\end{figure*}

We presented a new dataset for fine-grained road understanding
containing \num{11902} non-overlapping overhead images, associated
road attributes, and historical traffic data. \figref{coverage} shows
the spatial coverage of our dataset, with yellow, blue, and magenta
corresponding to the location of training, testing, and validation
images, respectively. \figref{data_supp} shows example images from the
dataset along with the road mask and a speed mask rendered using a
random time (road segments buffered to two meter half width).

\begin{figure}
    \centering
    \includegraphics[width=1\linewidth]{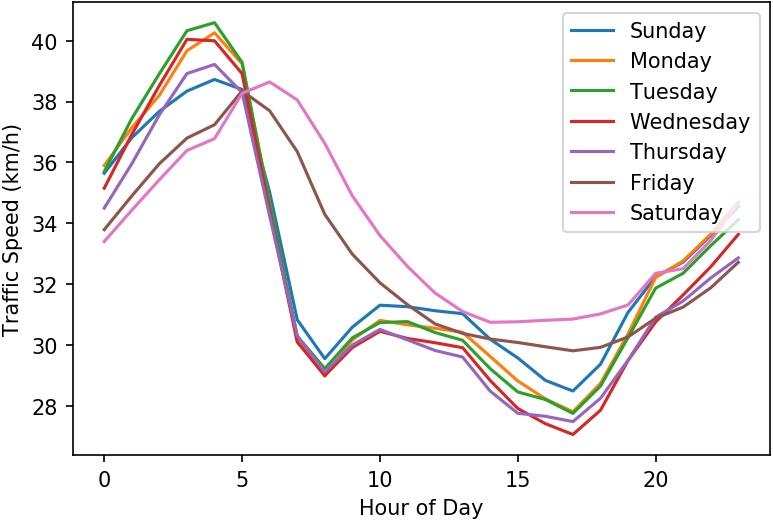}
    \caption{Average road segment speed versus time.}
    \label{fig:speed_vs_time}
\end{figure}

\begin{figure}
    \centering
    \includegraphics[width=1\linewidth]{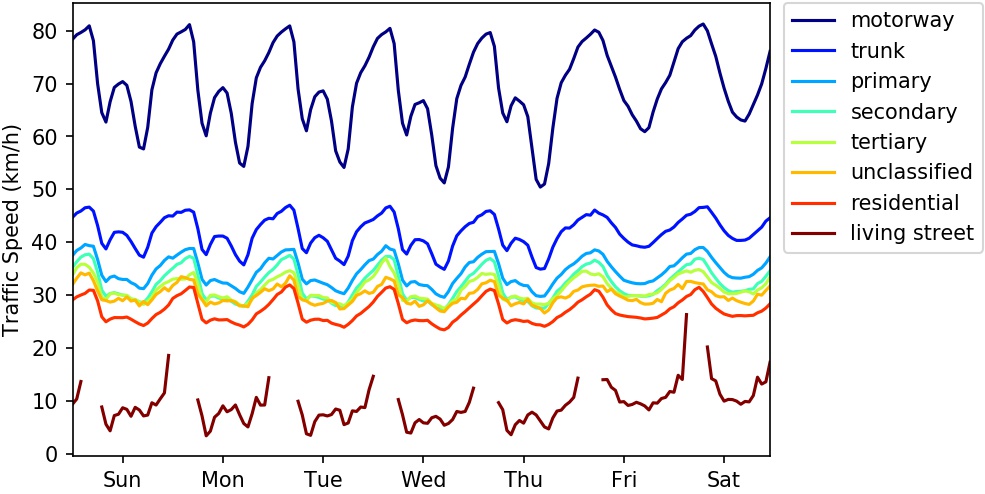}
    \caption{Average road segment speed versus time, where each road
    is categorized by its OpenStreetMap road type classification.}
    \label{fig:type_vs_time}
\end{figure}

We also report some statistics of the underlying historical traffic
speed data. Note that these numbers are computed on the provided
aggregated traffic speed data, which is then further aggregated by day
of week and hour of day. The average road segment speed in NYC for
2018 (averaging road segments over time first, then averaging across
segments) is approximately \num{30.47} km/h ($\sigma = 11.99$). In
\figref{speed_vs_time}, we visualize the average road segment speed
versus time. Additionally, \figref{type_vs_time} shows the average
speed according to OpenStreetMap's road type classification.

\section{Extended Evaluation}

In the main document, we presented an evaluation for traffic speed
estimation where each image in the test set was associated with a
random time to represent the ground truth speed data. These image/time
pairs were then fixed for all methods evaluated. Here we explore a
macro evaluation, where we consider the relationship between
performance and time. \tblref{macro} shows the results of this
experiment for a subset of times. When computing metrics, we select
all images in the test set which have at least one segment with
empirical traffic speed data at the time of interest. Then, we compute
metrics for each time and average the result (treating all times
equally). For this experiment, we consider Monday and Saturday and
selected the following hours of the day: 12am, 4am, 8am, 12pm, 5pm,
and 8pm. Consistent with our earlier results, our method that
integrates location and time outperforms an image-only baseline.

\begin{table}[!t]
  \centering
  \caption{Quantitative evaluation of traffic speed estimation (RMSE).}
  \begin{tabular}{@{}lccc@{}}
    \toprule
    & {\em image} & {\em Ours (uniform)} & {\em Ours} \\
    \midrule
    Monday (4am)     & 15.82 & 13.31 & \textbf{13.24} \\
    Monday (12pm)    & 10.96 & 10.59 & \textbf{10.41} \\
    Saturday (5pm)   & 10.65 & 10.39 & \textbf{10.36} \\
    Saturday (8pm)   & 10.43 & 10.32 & \textbf{10.27} \\
    \midrule
    Overall          & 11.903 & 11.145 & \textbf{11.134} \\
    \bottomrule
  \end{tabular}
  \label{tbl:macro}
\end{table}

\subsection{Impact of Angle-Dependent Speeds}

In our approach, we estimate angle-dependent speeds as opposed to
predicting a single speed per location. The intuition behind this idea
is that speed tends to depend on direction, e.g., a bridge that
crosses over a highway. Following the above evaluation scheme, we
compare our strategy of making angle-dependent speed predictions to a
baseline that instead uses uniform weights. In other words, we replace
the orientation-weighted average with an equal-weighted average. The
results are shown in \tblref{macro}. Our full approach outperforms the
uniform variant.

\begin{figure*}
  \centering
  \setlength\tabcolsep{1.5pt}
  \begin{tabular}{ccc}
    Shortest Distance  & Travel Time (Monday at 4am) & Travel Time (Monday at 8am) \\    
    \frame{\includegraphics[trim={2cm, 2cm, 2cm, 2cm},clip,width=.32\linewidth]{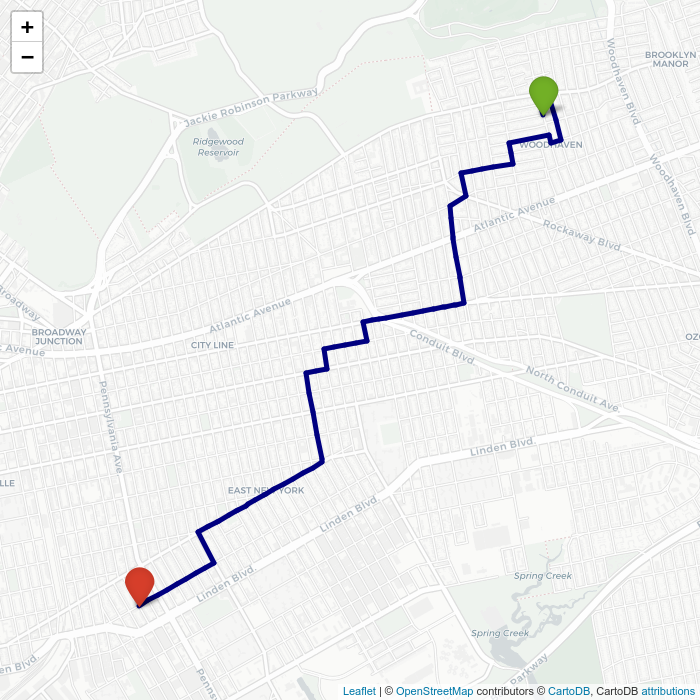}} &
    \frame{\includegraphics[trim={2cm, 2cm, 2cm, 2cm},clip,width=.32\linewidth]{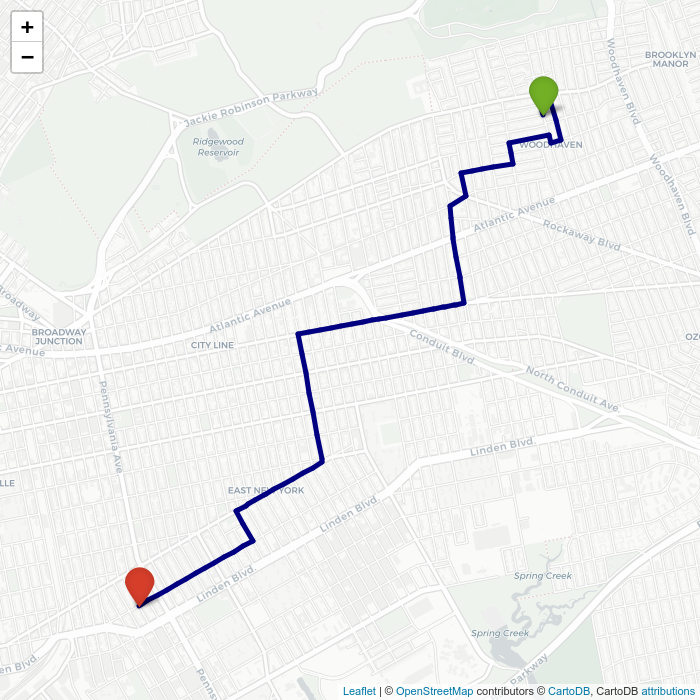}} &
    \frame{\includegraphics[trim={1.4cm, 2cm, 2.6cm, 2cm},clip,width=.32\linewidth]{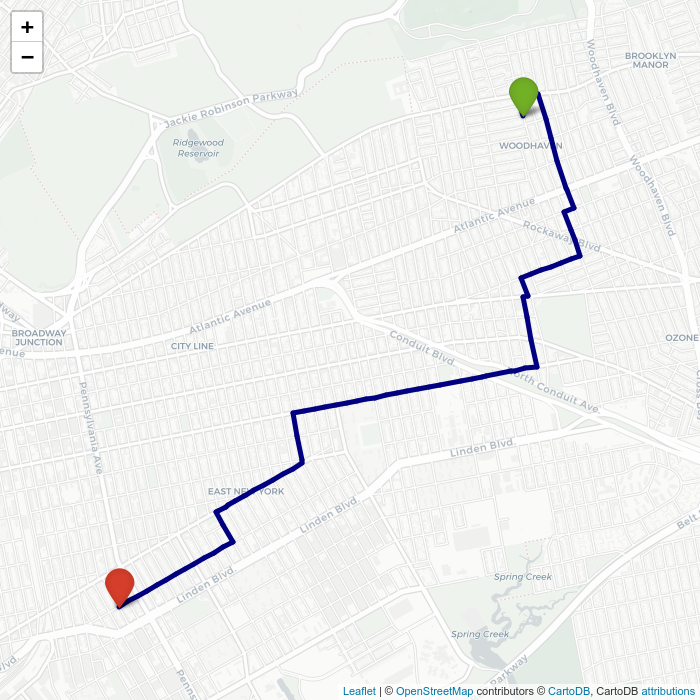}} \\
  \end{tabular}
  
  \caption{Using dynamic traffic speed predictions to augment route
  generation. The routes correspond to (left) the shortest path in
  terms of total length, (middle) travel time on a Monday at 4am, and
  (right) travel time on a Monday at 8am. Note that for the travel
  time routes, edge weights are represented by traversal times and
  computed using the length of the road segment and corresponding
  traffic speed estimated by our approach.}
  \label{fig:routing}

\end{figure*}

\section{Application: Augmenting Routing Engines}

Our method can be applied to generate optimal travel routes that take
into account traffic speeds at different times. For this experiment,
we use the OSMnx library~\cite{boeing2017osmnx} to represent the
underlying road topology. We compute the traversal time of each edge
based on the road segment length and our speed estimate.
\figref{routing} shows the results of this experiment for a route in
Queens, New York. \figref{routing} (left) shows the route
corresponding to shortest overall distance. \figref{routing} (middle)
shows the route corresponding to the shortest travel time on Monday at
4am. \figref{routing} (right) shows the route corresponding to the
shortest travel time on Monday at 8am.

\section{Additional Results}

\begin{figure}
    \centering
    \includegraphics[width=1\linewidth]{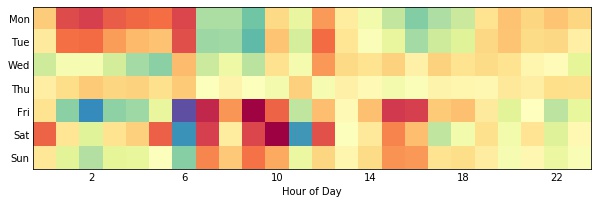}
    \caption{Visualizing the learned time embedding.}
    \label{fig:time}
\end{figure}

\paragraph{Visualizing the Time Embedding} In \figref{time} we
visualize the learned time embedding. To form this image, we take the
three dimensional embedding for each dimension of time (day of shape
$7 \times 3$ and hour of shape $24 \times 3$) and form a false color
image of shape $7 \times 24 \times 3$ by broadcasting; then average
across channels. As observed, the learned embeddings are clearly
capturing traffic speed patterns related to time. For example, on
Friday and Saturday the embedding reflects much larger values (red)
around the middle of the day as opposed to other days of the week.
This makes sense as many people work half days on Friday and leave
work early or take the family shopping on Saturday. This result also
agrees with the temporal patterns for traffic speed shown in
\figref{speed_vs_time}.

\begin{figure}
    \centering
    \setlength\tabcolsep{1pt}
    \begin{tabular}{cccc}
        Image & Target & Prediction & Error \\
        \includegraphics[width=.24\linewidth]{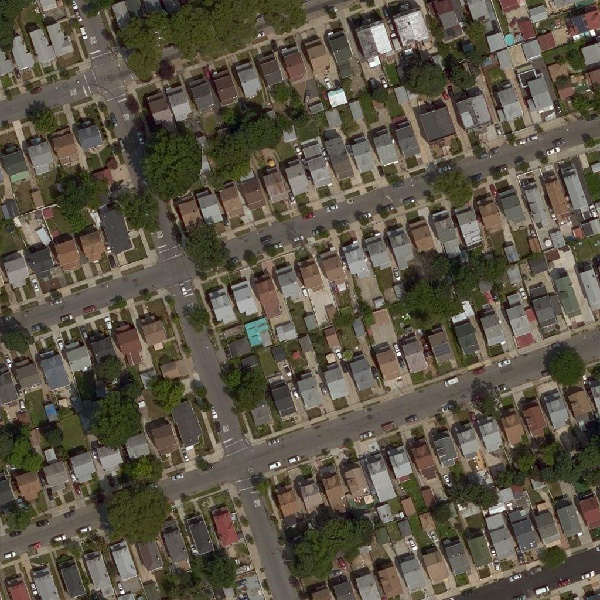} &
        \includegraphics[width=.24\linewidth]{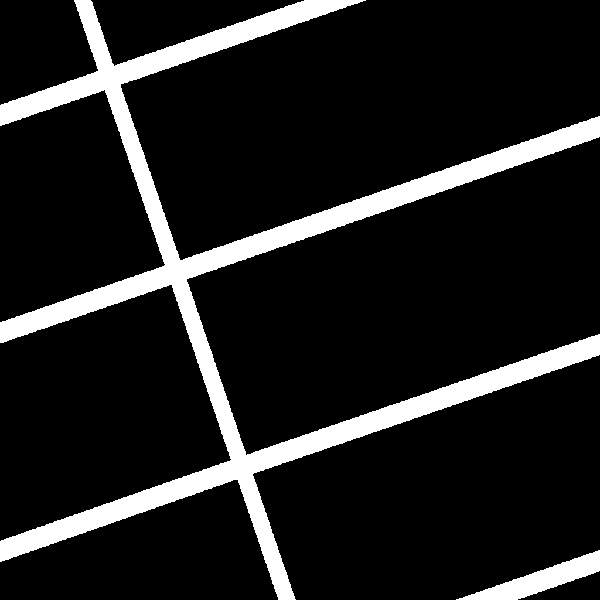} &
        \includegraphics[width=.24\linewidth]{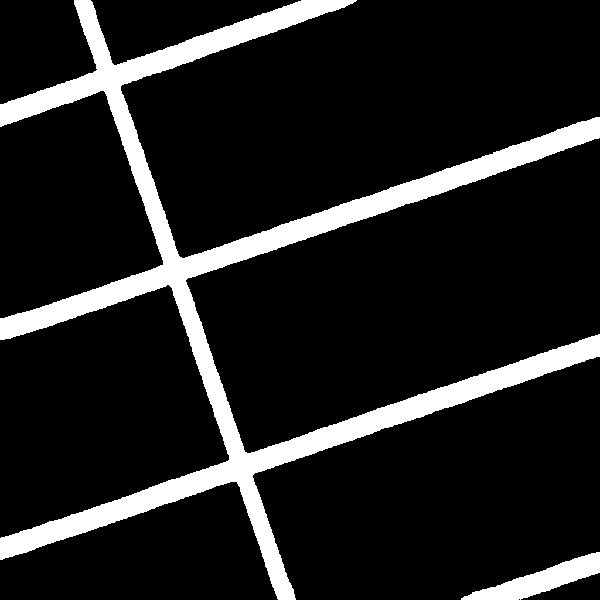} &
        \includegraphics[width=.24\linewidth]{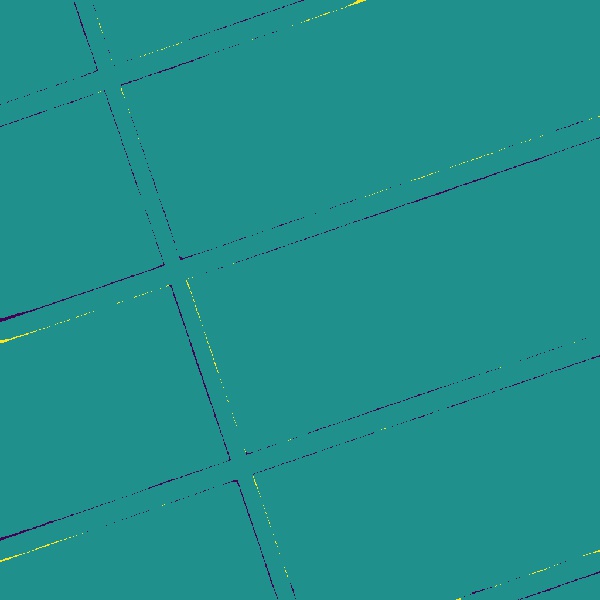} \\
        
        \includegraphics[width=.24\linewidth]{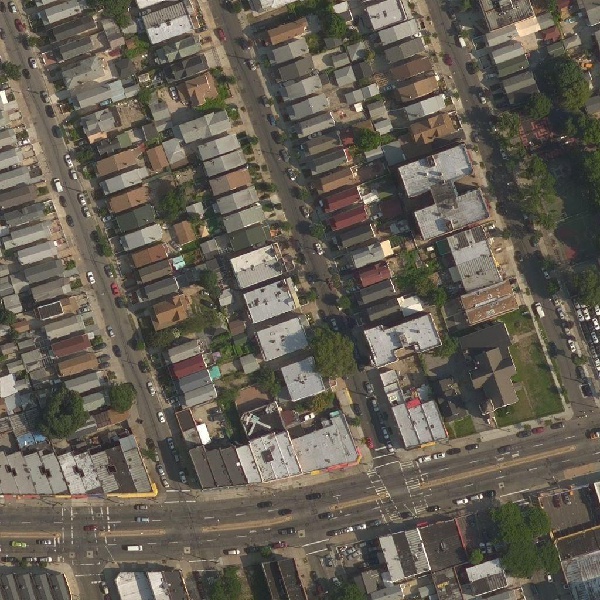} &
        \includegraphics[width=.24\linewidth]{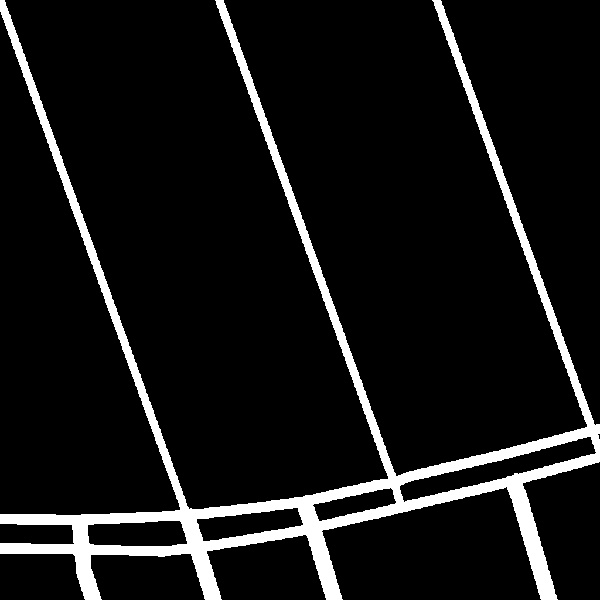} &
        \includegraphics[width=.24\linewidth]{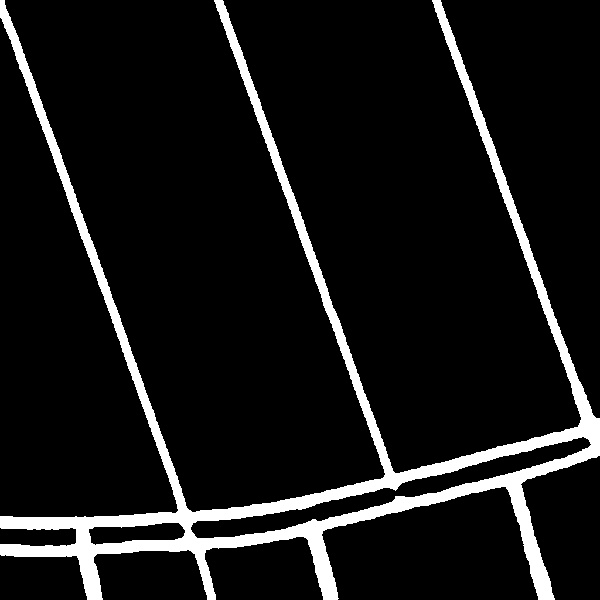} &
        \includegraphics[width=.24\linewidth]{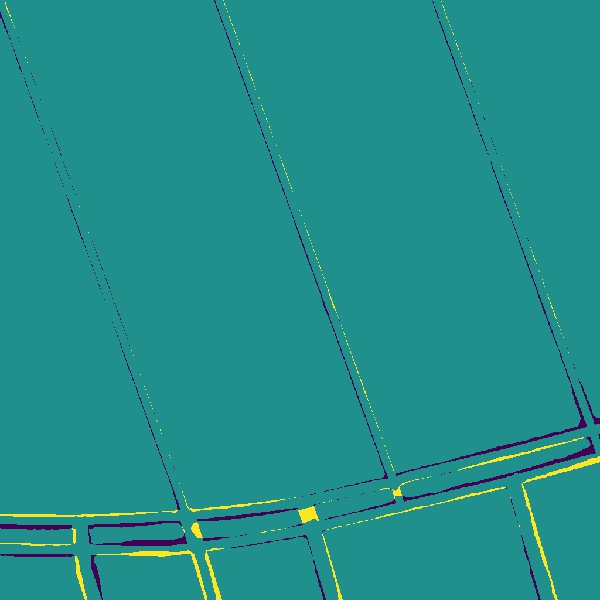} \\
        
        \includegraphics[width=.24\linewidth]{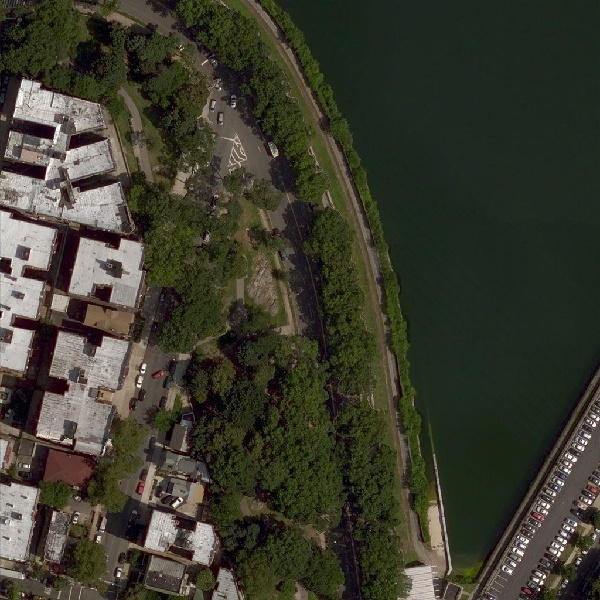} &
        \includegraphics[width=.24\linewidth]{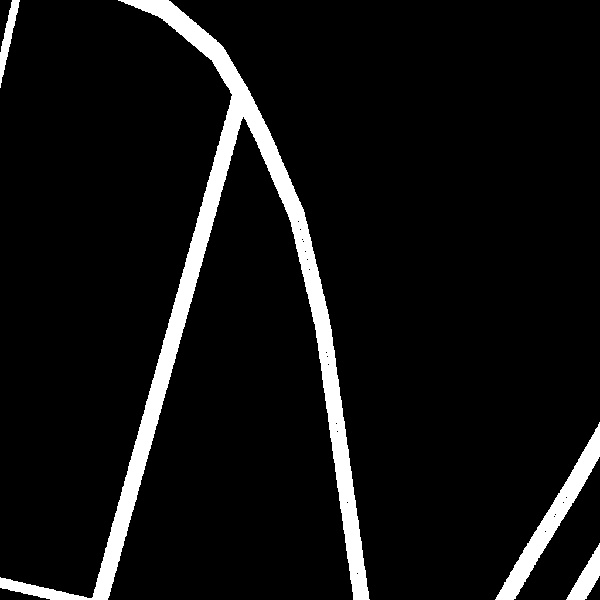} &
        \includegraphics[width=.24\linewidth]{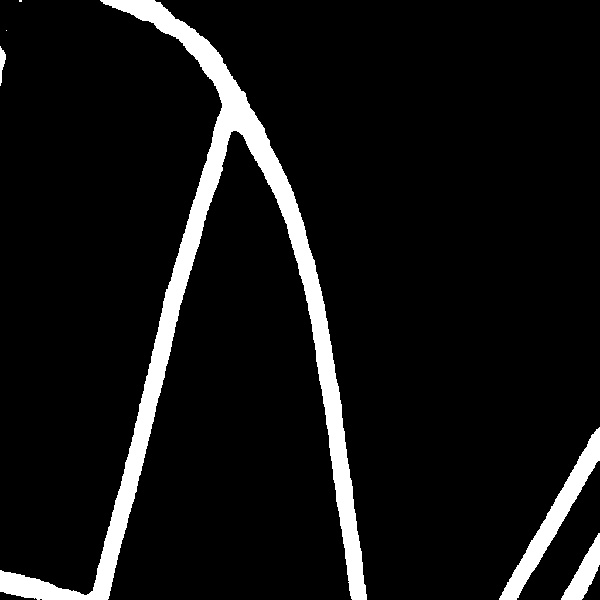} &
        \includegraphics[width=.24\linewidth]{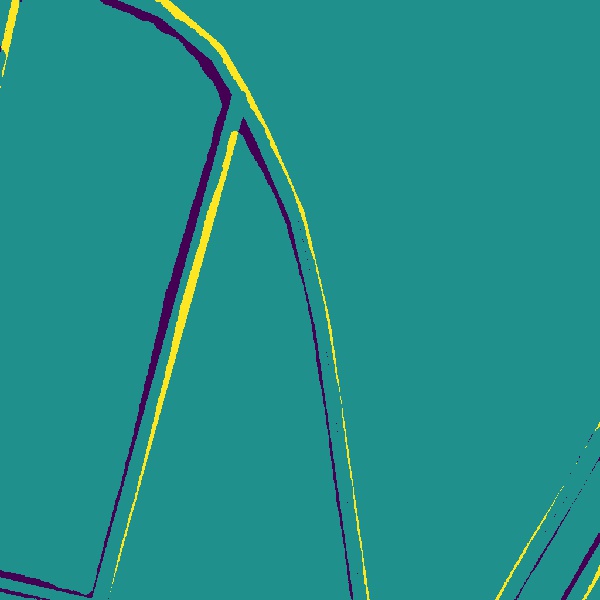} \\
    \end{tabular}
    
    \caption{Qualitative results from road segmentation. The error
    image (right) shows false positives (negatives) color coded as
    purple (yellow).}

  \label{fig:road}
\end{figure}

\begin{figure*}
    \centering
    \includegraphics[width=.41\linewidth]{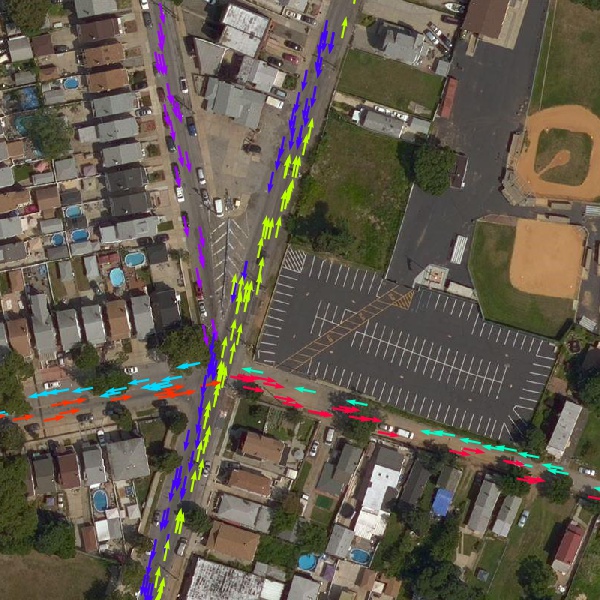}\\
    \includegraphics[width=.41\linewidth]{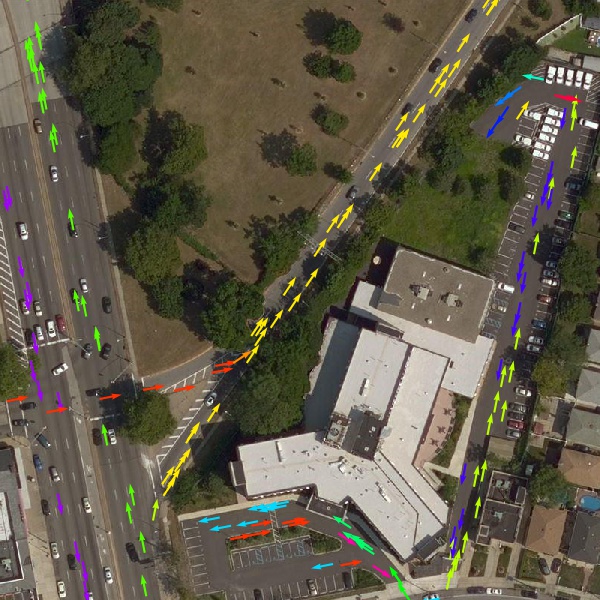}\\
    \includegraphics[width=.41\linewidth]{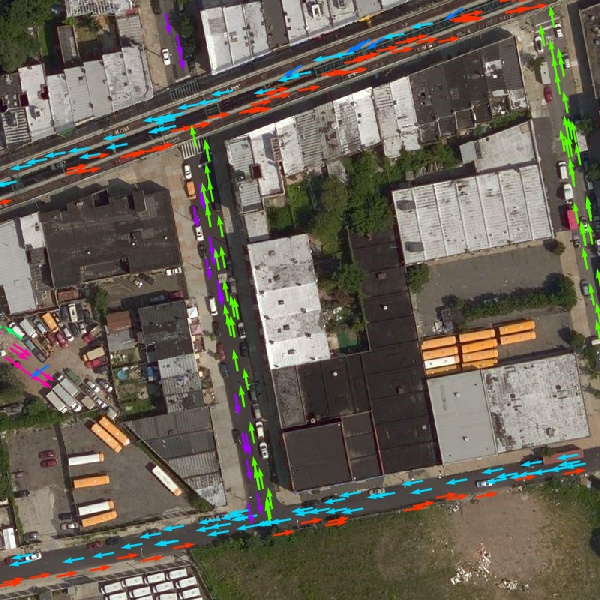}
    \caption{Using our approach to estimate directions of travel
    (visualized as flow fields).}
    \label{fig:flow}
\end{figure*}

\paragraph{Qualitative Results} \figref{road} shows several example
images alongside the ground-truth road mask and the prediction from
our approach. Similarly, \figref{flow} shows additional examples for
orientation estimation, visualized as a flow field for a subset of
points.

\end{document}